\documentclass[10pt,journal,compsoc]{IEEEtran}

\usepackage{amsmath}
\usepackage{amssymb}
\usepackage{etoolbox}
\usepackage[normalem]{ulem}
\usepackage{psfrag,epsf}
\usepackage{enumerate}
\usepackage{indentfirst}
\usepackage{enumitem}
\usepackage[numbers]{natbib}
\usepackage{mathalfa}
\usepackage{subfigure}
\usepackage{easyReview}
\usepackage{prodint}
\usepackage{comment}
\usepackage{algorithm}
\usepackage{algcompatible}
\algnewcommand\INPUT{\item[\textbf{Input:}]}%
\algnewcommand\OUTPUT{\item[\textbf{Output:}]}%
\usepackage{url} 
\usepackage{bm}
\usepackage[utf8]{inputenc}
\usepackage[english]{babel}
\usepackage[colorlinks,citecolor=blue,urlcolor=blue]{hyperref}
\usepackage{longtable,booktabs}
\usepackage{mathtools}
\usepackage{color, colortbl}
\definecolor{Gray}{gray}{0.9}
\definecolor{Gray2}{gray}{0.7}
\usepackage{rotating}
\usepackage{tikz}
\usetikzlibrary{trees,shapes.geometric,shapes.misc,arrows.meta,chains, arrows,positioning,shapes,shadows,shapes.multipart}

\def\b0{\mathbf{0}}

\def\bD{\mathbf{D}}

\def\bR{\mathbf{R}}

\def\bx{\mathbf{x}}

\def\bX{\mathbf{X}}
\def\bY{\mathbf{Y}}

\def\rT{\mathrm{T}}

\ifCLASSINFOpdf
\else
\fi
\begin{document}
\title{An Overview of Healthcare Data Analytics with Applications to the COVID-19 Pandemic} 

\author{Zhe~Fei, 
        Yevgen~Ryeznik, 
        Oleksandr~Sverdlov, 
        Chee~Wei~Tan, 
        and~Weng~Kee~Wong
\IEEEcompsocitemizethanks{\IEEEcompsocthanksitem Z. Fei and W.K. Wong are with the Department of Biostatistics, UCLA, Los Angeles CA, 90095 USA\protect\\
E-mail: feiz@ucla.edu, wkwong@ucla.edu.
\IEEEcompsocthanksitem Y. Ryeznik is with the AstraZeneca, Gothenburg, Sweden. e-mail: yevgen.ryeznik@gmail.com.
\IEEEcompsocthanksitem O. Sverdlov is with Novartis Pharmaceuticals Corporation, New Jersey, USA. E-mail: alsverdlov@gmail.com.
\IEEEcompsocthanksitem C.W. Tan is with the Department of Computer Science, City University of Hong Kong, Hong Kong. E-mail: cheewtan@cityu.edu.hk.
}
\thanks{This work was funded in part by a grant from the Hong Kong ITF Project No. ITS/188/20, UGC Teaching Award Project No. 6989041 and an Institute for Pure and Applied Mathematics (IPAM) Senior Fellowship.  Manuscript received Oct., 2020; revised June, 2021; accepted July 2021.}}



\markboth{IEEE Transactions on big data,~Vol.~XX, No.~X, Aug 2021}%
{Fei \MakeLowercase{\textit{et al.}}: Advances in Data Science}
%



\IEEEtitleabstractindextext{%
\begin{abstract}
In the era of big data, standard analysis tools may be inadequate for making inference and there is a growing need for more efficient and innovative ways to collect,  process, analyze and interpret the massive and complex data. We provide an overview of challenges in big data problems and describe how innovative analytical methods, machine learning tools and metaheuristics can tackle  general healthcare problems with a focus on the current pandemic. In particular, we give applications of  modern  digital technology, statistical methods,data platforms and data integration systems to  improve diagnosis and treatment of diseases in clinical research and novel epidemiologic tools to tackle infection source  problems, such as finding Patient Zero in the spread of epidemics.  We make the case that analyzing and interpreting big data is a very challenging task that requires a  multi-disciplinary effort to continuously create more effective methodologies  and  powerful tools to transfer data information into knowledge that enables informed decision making.
\end{abstract}

}

\maketitle

\IEEEdisplaynontitleabstractindextext

%
\IEEEpeerreviewmaketitle

\section{\bf Introduction}
The ongoing global COVID-19 pandemic  presents to us daily, if not hourly, updated sets of massive and messy data from all over the world and a continuous series of challenging research questions in multiple areas.  They include issues in data management, data analysis and interpretation and public health policies development that range from disease prevention and management to social concerns about mental health of the general public due to prolonged quarantine periods and restrictions in personal freedom. Massive and complex data, structured or unstructured,  are now becoming available in practically all disciplines, particularly in health data \citep{bigdata}.  The volume and speed at which {massive} data become available can make conventional methods for analyzing them less efficient or inappropriate. 

Identifying quantities of interest and making meaningful summary statistics for trends, patterns and relationships/associations among the  the different types of variables become an overwhelming task because of the huge number of variables in the data sets.  Even visualizing such data sets correctly can be extremely challenging and easily subject to manipulation and mis-interpretation \citep{highschool}. Data science  is a recognized discipline that develops analytic tools to effectively manage, analyze and interpret big data of various types. The field is rapidly evolving and fuels constant discussion in various disciplines; for instance in statistics and machine learning, recent perspectives on data science can be found in \citep{smirnova2018,jrssa1} . Its expanding important role to uncover vital insights in big data is now instrumental in many large-scale applications such as healthcare data analytics---the topic of focus in this paper.

Frequently, research questions are formulated  into various types of inferential problems, that likely include  studying associations among the massive number of different types of the heterogeneous variables, identifying risk factors for selected outcomes and  predicting future trends.  A distinguishing feature of the current pandemic is that it requires urgent, innovative and effective  analytic tools to obtain timely information that enables public health leaders to  make data-based strategic  decisions with confidence.  This paper discusses an overview of  innovative analytic approaches for tackling COVID-19 related problems using modern digital technologies, innovative statistical methodology for accurate inference from  big and complex data sets,  analytic  epidemiological  tools to track and control disease progression, and  state-of-the-art algorithms to compute and search for optimal strategies. The collective tools described herein are not limited to tackling COVID-19 problems and can be applied to solve other types of medical problems, and beyond.  For example, the epidemiological tools in Section \ref{sec4} can be \textcolor{black}{directly} modified to detect fraud and news leakage  or monitor and identify the key sources of fake news.  

\begin{table*}
    \centering
    \caption{COVID-19 related applications in the paper.\label{tab1}}
    \begin{tabular}{p{0.22\textwidth}p{0.28\textwidth}p{0.28\textwidth}p{0.12\textwidth}}
         Topic &  Methods & Impact & Reference \\
         \hline
         Digital Health & Wearable devices/technologies, digital therapeutics (DTx)	& Remote health monitoring and treatment delivery &	Sec. \ref{subsec21}, \ref{subsec22} \\
         Deep Learning & CNN, RNN & Diagnosis and classification of COVID-19 cases & Sec. \ref{subsec23} \\
         Scientific~Machine~Learning & ODE-based SIR model; Safe Blues & Modeling of limited data at the beginning of the pandemic; contact tracing app	& Sec. \ref{subsec24} \\
         High Dimensional Inference & LDPE; Debiased LASSO; SSHDI & Statistical inference for future COVID-19 related genomics data & Sec. \ref{sec3.4}\\ 
         Computational Epidemiology	& Graph-theoretic statistical inference; GNN	& Patient Zero search; Infodemic risk management, fake news	& Sec. \ref{sec4} \\
         Metaheuristics	& Nature-inspired algorithms: ICA; DE; PSO; CSO	& Disease trend prediction, spread monitoring, real-time prediction 	& Sec. \ref{sec5.1} \\
         \hline
         \multicolumn{4}{p{0.95\textwidth}}{
         Abbreviations: CNN = Convolutional Neural Networks; RNN = Recurrent Neural Networks;  ODE = Ordinary Differential Equations; SIR= Susceptible, Infectious, Recovered; LDPE = Low Dimensional Projection Estimator; LASSO = Least Absolute Shrinkage and Selection Operator; SSHDI = Split and Smoothing for High Dimensional Inference; GNN = Graph Neural Networks; ICA = Imperialist Competitive Algorithm; DE = Differential Evolution; PSO = Particle Swarm Optimization; CSO = Competitive Swarm Optimizer}
    \end{tabular}
\end{table*}

In the next few sections, we give an overview of the latest advances in data science with a focus on digital technologies for clinical research, statistical inference for big data and epidemiology. Neural networks, machine learning and metaheuristics are important tools in artificial intelligence and their relevance to solving COVID-19 problems is also mentioned.  Table \ref{tab1} identifies and  summarizes specific applications to COVID-19 problems in this paper.

Section \ref{sec2} discusses data science-based approaches to address important clinical research questions. We consider two examples. The first concerns designing a clinical study with exploratory tools, digital technologies and biomarkers to characterize depression,  which has seen a spike in the current pandemic \cite{ettman2020}. We discuss both conventional and innovative ways to analyze large volume, high frequency data in this setting, and emphasize the importance of careful formulation of research questions to address the scientific goals of the study. Our second example showcases a new branch of artificial intelligence research called Scientific Machine Learning (SciML).  SciML enriches mathematical models and facilitates the use of data-driven machine learning techniques to improve the quality of model-based prediction. As an application, we describe some virtues of the SciML approach to improve efficiency of  global COVID-19 quarantine policies.

Section \ref{sec3} describes some state-of-the-art methodologies and algorithms for making inference for big data that can provide some new insights into statistical inference with regression models and many more predictors than samples, referred to as ``high dimensional inference.'' High dimensional inference has broad applications and particular relevance to the joint modeling of features in large data sets, as illustrated by our application using a cancer genomic data set.  We demonstrate  the utilities of a newly proposed method, where we identify important gene pathways for early diagnosis of a disease via  finding significant   predictors among hundreds or more of them. The described techniques can be more broadly applied to other areas and data sets, such as the massive  COVID-19 data sets continuously generated from the Johns Hopkins University depository,  and the inference sought can be estimating or updating estimates of the various risk factors of COVID-19 and accurately  identifying significant predictors from  a large pool.

Section \ref{sec4} reviews applications of large-scale computational epidemiology optimization problems such as infection source detection (e.g., searching for Patient Zero) and its related Infodemic management due to the COVID-19 Pandemic. Solutions to these computational epidemiology optimization problems can provide health authorities with {\it digital contact tracing} to trace the social contacts of an infected person and searching for the outbreak origin \cite{contacttracing1,contacttracing2} or to formulate appropriate healthcare policies in the face of misinformation \cite{infodemic}. 

Section \ref{sec5} provides an overview of  increasing use of metaheuristics in various disciplines, including recent applications of metaheuristics to tackle multiobjective optimization problems related to COVID-19. The paper concludes in Section \ref{sec6} by emphasizing on the importance of multidisciplinary research, and the continuing central role of statistical thinking in the era of big data. 


\section{\bf{Data Science in Modern Clinical Research}\label{sec2}}
\subsection{{\bf Opportunities and Challenges}\label{subsec21}}
The 21st century biomedical research arena has and continues to benefit from the increasing computational power, innovative technologies and availability of big data. The term ``big'' refers to several characteristics of the data that are also referred to as the ``V's of big data''\citep{bigdata01, bigdata02, bigdata03}. There are at least six V's: volume, variety, velocity, veracity, variability (and complexity), and value. \textit{Volume} refers to the magnitude (amount) of data, which depends on the technology development and is continuously increasing. \textit{Variety} refers to the structural heterogeneity of data sets, e.g. structured, semi-structured, and unstructured data. \textit{Velocity} refers to the rate of data generation and processing, which keeps growing with the advances in technology. \textit{Veracity} corresponds to uncertainty and unreliability in the data sources  due to subjectivity of human opinions or in social media. \textit{Variability} (and complexity) refers to the variation in the data flow rates (which can have peaks and troughs) and complexity in data processing due to heterogeneity of the data sources. \textit{Value} corresponds to the benefit that data adds to the enterprise, e.g. increased revenue, decreased operational costs, higher customer satisfaction, etc.

Big data brings tremendous opportunities and new multidisciplinary challenges for clinical research. Here are just a few examples of big data sources in this context.

\begin{itemize}
\item {\bf \emph{Real World Evidence (RWE)}}:  In the 20th century, the randomized controlled trials (RCTs) were established as the gold standard of evidence-based research for evaluating safety and efficacy of new treatment interventions \cite{harrington2000rct}. However, RCTs may be long, expensive, and difficult to conduct. Alternative sources of clinically important data include electronic health records (EHRs), population based registries, and some other real-world data that can supplement and generalize the evidence from RCTs. Development of big electronic databases has enabled collection and integration of such RWE; yet, it is still complex, multi-dimensional, and lacking clear structure. For instance, many drug prescriptions were hand-written and later scanned and saved electronically. How can one extract and then classify this important information? An increasingly promising approach is the \emph{natural language processing} (NLP) \citep{or11}. Organization of the NLP to ensure proper data collection, cleaning, restructuring, and getting it to a point when it can undergo a meaningful analysis is both a challenge and an opportunity.

\item {\bf \emph{Medical Imaging Technology}} 
: Many clinical trials utilize objectively measured biomarkers that capture disease progression over time and provide measurement of treatment effects. The \emph{magnetic resonance imaging} (MRI) has now been widely used in clinical research in Alzheimer's disease \citep{cash2014alzheimer}, multiple sclerosis \citep{or2}, cancer \citep{or9}, etc. Analysis and interpretation of MRI data requires high medical expertise and judgement, and it is also very time consuming and expensive. Automating this process could provide a more objective and less costly way of extracting important medical information. 
\emph{Convolutional neural networks} (CNN) is a class of deep learning methods that can be useful for analyzing MRI data, to produce objective, high quality outcome measures \citep{or1}. This can potentially improve signal-to-noise ratio and increase the efficiency of clinical trials.

\item {\bf \emph{Digital Endpoints}}: Novel sensors and wearable technologies (e.g., smart watches) have enabled collection of terabytes of individual health information, such as physical activity, vital signs, quality of daily living, etc. These data can be collected with high frequency over extended time periods, and provide means to identify serious medical problems (e.g. heart abnormalities that could lead to a heart attack). Big data generated by wearable technologies can potentially reduce the need for clinical site visits and streamline the clinical trial research. This can be especially valuable during a global pandemic such as COVID-19, when hospitals and clinical trial research units are overwhelmed and patients are often unable to keep their scheduled in-clinic visits due to quarantine restrictions. However, this promise comes with the need for careful data collection, processing, and development of valid digital endpoints \citep{barbak}.

\item {\bf \emph{Digital Therapeutics}} (\textbf{\emph{DTx}}): A true hallmark of the 21st century medicine is the development of DTx -- evidence based therapeutic interventions driven by high quality software programs to prevent, manage, or treat a medical disorder or disease \citep{or15}. The unmet medical need addressed by DTx is very diverse. As an example, consider the \emph{precision dosing} paradigm in management of different chronic diseases \citep{polasek2018precision}. \emph{Closed-loop systems} that automatically determine optimal time of blood sampling and perform calculation of the optimal dose and timing of dose delivery provide means for individualizing treatment to patient. For instance, there is emerging clinical trial evidence that closed-loop insulin delivery systems can improve glucose control in patients with type 1 diabetes \citep{benhamou2019lancet}, and these systems are expected to become standard soon. A potential virtue of DTx is magnified during the COVID-19 crisis, as DTx products can deliver safe and effective care remotely (\url{https://www.fda.gov/media/136939/download}).
\end{itemize}

There is clear benefit of analyzing large clinical data sets but there are constant debates on the analytic approaches \citep{or3}. In the next subsection, we present an example of a clinical study evaluating different digital technologies, where both big data and traditional clinical data are collected. We argue that most scientific and statistical principles still apply in such experiments, while some novel data science and machine learning techniques can nicely supplement the more traditional and established approaches.

\subsection{An Example of a Clinical Study Evaluating Digital Technologies\label{subsec22}}

Depression is a burdensome mental health disorder that often goes undetected and untreated \citep{or6}. Symptoms of depression are multi-dimensional and affect emotions, thoughts, behavior, and physical domains. 
There is a strong need to develop effective  methods to  diagnose depression and perform efficient monitoring of patients with this condition.

Conventional measures of depression-related symptoms are paper-and-pen outcome assessments, such as the 
Hamilton Depression Rating Scale (HDRS) \citep{or8} and the Montgomery-Åsberg Depression Rating Scale (MADRS) \citep{or12}. While these measures are well-established, they are subject to rater bias, may lack clinical relevance and exhibit high variability, which translates into the need for large clinical trials to detect clinically relevant treatment differences. On the other hand, digital technologies 
have the potential to provide more objective and precise tools to detect depression-related symptoms; yet, these technologies require careful assessment and validation in clinical studies before they can be broadly implemented.

Our example is a single-site, cross-sectional, non-interventional study of novel exploratory tools, digital technologies and biomarkers to characterize depression. 
 The study results have been  published in \citep{sverdlov2021}. Here we discuss some important aspects of the study design and illustrate the thinking process and the logic behind selection of appropriate data analysis tools.

 The study evaluated 40 subjects (20 patients with major depressive disorder (MDD) and 20 healthy volunteers). There were three in-clinic visits at days 1, 7, and 14. At each visit, study subjects underwent a series of assessments, both conventional (e.g. MADRS) and novel digital technologies. In addition, between visits there was at-home collection of data through mobile apps. The study objectives were three-fold:
\begin{itemize}
\item	To assess feasibility of use of digital technologies.
\item	To assess utility of these technologies as diagnostic tools (classification of subjects, MDD vs. healthy).
\item	To explore a potential of using digital biomarkers as predictors of the conventional measures (MADRS).
\end{itemize}
 Due to the small sample size, this study was exploratory in nature. It provided only preliminary evidence on virtues of digital technologies, which has to be further confirmed in larger studies. Overall, seven digital technologies were evaluated. These can be broadly categorized as mobile apps that provided real-world data and tests that were performed in-clinic.

One mobile app provided an interactive tool for high-frequency assessments of cognition and mood over the course of the study. Another mobile app was a passive behavioral monitor that integrated smartphone data related to the user's social acts and patterns, e.g. phone calls, social media use, travel, etc. A third app was a platform to perform voice recordings to obtain vocal biomarkers that contain important information on depression-related symptoms. 

The in-clinic digital technologies included a neuropsychological test battery; an eye motor tracking system that captures information across multiple domains of mood, cognition, and behavior; an electroencephalogram (EEG)-based technology to analyze the brain network  activity; and a task quantifying bias
in emotion perception.

 The study data was very diverse and varied in structure and complexity. For instance, for a behavioral monitor app, a list of 85 features  was derived and later scrutinized to 10 most important features per subject.
These features represented various summary measures of social functioning, e.g. total duration of all communication events, entropy of the usage time of social media apps, number of places visited, etc. Therefore, despite high volume and high frequency of the raw data, for each subject we obtained a vector of numeric summaries. By contrast, for the interactive app that provided measurements of cognition and mood, we {acquired} longitudinal data per subject: a cognitive score and a mood score were calculated each time the subject engaged with the app.
For a neuropsychological test battery, the data were acquired at each in-clinic visit, 60, 160, and 230 minutes after admission, and then averages across three time points were taken, and a total of 42 features were derived per subject per visit. 
 The vocal biomarkers were derived by applying signal processing algorithms on various time windows within a sample of speech, for a total of 72 features per subject.

The data from digital technologies was combined with demographic and clinical questionnaire data for analysis, which included exploratory data visualizations and different supervised learning techniques. 
 For instance, we performed classification analysis to predict the class of each subject, MDD or healthy, using logistic regression.
Multiple linear regression  was used to model and predict MADRS total score as a function of digital biomarkers. 
 Our analysis was organized by technology -- to understand utility of each technology and identify digital biomarkers that add most value. In many cases we were able to develop simple, parsimonious models with reasonably high diagnostic accuracy and potential to predict standard clinical outcome in depression \citep{sverdlov2021}.

 One major lesson learned from this study is that while many novel digital technologies generated large-volume, high-frequency data, the majority of good clinical and statistical research principles were still applicable in this setting. The common techniques of data analysis such as classification and regression could handle many types of the data described.  However, analysis of some data types, such as speech samples, require more advanced machine learning techniques. In the next section, we describe approaches based on neural networks that can be potentially useful for analysis of such complex data, and the computational tools for implementing neural networks.

\subsection{Neural Networks and Distributed Computing for Clinical Research\label{subsec23}}
Recent developments in computational technologies and increasing computational powers allow exploration and analysis of data with very complex structure. The sound data and the video data might have a potential application in disease diagnostics \citep{bourke,ozkanca}. One may consider this type of data as a time series with a non-numeric outcome. A video object can be viewed as a time series with picture outcomes, and a sound can be viewed as a time series of multidimensional vectors, or even graphs.  Analysis of this type of data requires both new statistical and numerical approaches and highly powerful computational software. A combination of algorithms based on neural networks and distributed computing seems to be a viable approach to perform such an analysis.

A neural network model is based on simplified assumptions of the human brain architecture. The model is highly parameterized and requires computationally intensive optimization for the parameters tune-up. The process of the parameters' optimization is called \textit{learning}. The simplest example of a neural network is a single neuron (perceptron) model \cite{or13}. In this model, a neuron accepts $n$  inputs  (usually, numerical), referred to as  \emph{covariates} in statistics terminology or \emph{features} in machine learning. Each input point $x_i$ has some positive weight $w_i$. Then, a weighed sum of input points is substituted as an argument of an activation function $f$, and the value obtained is an output of the model. A common choice is $f(x)=1/(1+e^{-x})$, a sigmoid function  with a \emph{bias term} $b$ and an output of a single neuron is 

$$
  y = f\left(b + \sum_{i = 1}^n{w_ix_i}\right).
$$

The perception model was proposed for a binary classification problem in 1958  \cite{or13}. The recent dramatic improvements in  computational power has resulted in more complex and effective neural network architectures, and they include \textit{deep learning} (DL), where the neural network contains many connected neurons and may have several hidden layers and several outputs. Every single neuron in the network can learn a simple input-output relationship. Then, all the neurons exchange the information learned to make the entire learning feasible. At a glance, it may seem that a neural network is a nonlinear approximator that is not much different from other approximation models. This may be true. For example, \cite{or4} proved that a network with a single hidden layer containing a finite number of neurons and a sigmoid activation function can approximate continuous functions on compact subsets of $\bR^n$. The theorem thus states that simple neural networks can represent a wide variety of nonlinear functions when given appropriate parameters. However, neural networks can be scaled, extended and generalized in a variety of ways: more hidden units in a layer, multiple hidden layers, weight sharing, innovative learning algorithms for massive data sets \citep{or5}.

There are two special classes of deep neural networks: \textit{convolutional neural networks} (CNNs) and \textit{recurrent neural networks} (RNNs). The CNNs' architecture is inspired by the structure of the animal visual cortex, which makes CNNs extremely useful in applications dealing with imaging data, particularly, in medical image analysis. Thus, the algorithms using CNNs for processing and analysis of computer tomography, CT Scans, and chest X-ray images have been found efficient for diagnostic and classification of COVID-19 cases \citep{cnn-covid01, cnn-covid02, cnn-covid03}. The RNNs are designed in such a way that connections between their nodes represent a directed graph along a temporal sequence, which allows to model dynamic behavior over time. The RNNs' architecture makes them capable of using their internal state as a memory to process the inputs of variable length and to model time series with more complex observations. This makes RNNs applicable for solving such problems as handwriting or speech recognition. Some of the recent examples of using RNNs to monitor the COVID-19 situation can be found in \citep{rnn-covid01, rnn-covid02}. A computational issue with classic RNNs is that when training it using back-propagation, the gradients which are back-propagated can ``vanish'' (i.e., go to zero) or "explode" (i.e., go to infinity). As a special case of RNNs, long short-term memory (LSTM) avoids the vanishing gradient problem by using recurrent gates called ``forget gates'' which allow gradients to flow backwards and unchanged. However, LSTM networks can still suffer from the exploding gradient problem \cite{gers2002learning,bayer2009evolving}.  Extensions of RNNs continue to develop; a recent example is the Transformer, which is a new model used primarily in the field of natural language processing (NLP), which, unlike RNNs, does not require that the sequential data be processed in order.

The procedure of training a neural network includes optimization of weights by minimizing some loss function, given a training data set. Usually, the loss measures how different are the outputs produced by a neural network and the true responses taken from a training data set. In most of the scenarios, optimization is performed using gradient-based methods. The backpropagation algorithm \cite{or7} is used for efficient calculation of gradients. While the gradient-based methods are good optimization techniques which work excellently for convex functions and low dimensional space, one may expect much better results with the particle swarm optimization (PSO) technique \citep{or10,or11}. However, there appears to be very few publications using PSO as a tool to train neural networks, and it certainly merits investigation.

 Table \ref{tab2} provides a short list of current computational tools  for implementing neural networks in practice.

\begin{table*}
    \centering
    \caption{Computational tools for neural networks and distributed computing.\label{tab2}}
    \begin{tabular}{p{0.12\textwidth}p{0.78\textwidth}}
         Tool &  Description \\
         \hline
         Tensorflow & An open software library for ML developed by Google. Accessible via \url{https://tensorflow.org}. The main API for working with the library is implemented for Python; there are also implementations for C Sharp, C++, Haskell, Java, Go, and Swift. \\
         \rule{0pt}{3ex} Caffe & Caffe stands for \textit{Convolutional Architecture for Fast Feature Embedding}. It is an open-source DL framework written in C++ with a Python interface. It was developed at the University of California, Berkeley and is accessible via \url{http://caffe.berkeleyvision.org/}. It supports various DL architectures for image classification and segmentation, as well as GPU- and CPU-based acceleration computational kernel libraries such as NVIDIA cuDNN and Intel MKL. \\
         \rule{0pt}{3ex} PyTorch & An open-source ML library used for applications such as computer vision and NLP. The supported programming languages include Python, C++, and CUDA. PyTorch has been primarily developed by Facebook's AI research group. At the end of March 2018, Caffe and PyTorch were merged. \\
         \rule{0pt}{3ex} Keras & An open-source NN library written in Python on top of TensorFlow and some other DL libraries. It was designed with a focus on being user-friendly, modular, and extensible, and it allows fast experimentation with DNN, as well as CNN and RNN. Accessible via \url{https://keras.io/}.\\
         \rule{0pt}{3ex} MapReduce \cite{or17} & Provides a framework for computing some sets of distributed tasks using a large number of computers (called ``nodes'') that make up a cluster. The scope of  MapReduce consists of two steps: Map and Reduce. At the Map step, one of the computers (called the master node) receives the input data of the task, splits it into parts, and transfers it to other computers (work nodes) for preliminary processing. At the Reduce step, the pre-processed data is collapsed. The main node receives responses from the working nodes and on their basis forms the result, i.e. the solution to the originally formulated problem. \\
         \rule{0pt}{3ex} Apache Spark & An open-source framework for implementing distributed processing of unstructured and weakly structured data.  It provides computations built around resilient distributed data sets (RDDs). Unlike MapReduce that operates with disk storage, Apache Spark uses RDDs for recursive processing in RAM, thereby enabling it to perform more efficiently for some classes of tasks. It supports high-level tools for SQL queries and structured data processing (Spark SQL), ML problems (MLlib), graph processing (GraphX), and stream processing of live data streaming (Spark Streaming). Apache Spark is a key platform for distributed DL; it allows embedding of TensorFlow, and other DL frameworks in Spark workflows, to build distributed DL applications. Accessible via \url{https://spark.apache.org/}. \\ 
         \rule{0pt}{3ex} Databricks & A platform from the Apache Spark creators that provides functionality for reproducible research. It supports several programming languages (e.g. Spark, SQL, Java, Scala, Python, R) and allows easy switching  between different languages within a project.  Accessible via \url{https://databricks.com}. \\ 
         \hline
         \multicolumn{2}{p{0.95\textwidth}}{
         Abbreviations: CNN=Convolutional Neural Networks; DL=Deep Learning; DNN=Deep Neural Networks;  ML=Machine Learning; NLP=Natural Language Processing; NN=Neural Networks;  RNN=Recurrent Neural Networks}
    \end{tabular}
\end{table*}

\subsection{Scientific Machine Learning\label{subsec24}}
Recently, a new branch of AI research has evolved at the edge of scientific computing and machine learning (ML)—Scientific Machine Learning (SciML, https://sciml.ai).  Scientific computing deals with mathematical models of real-life systems based on physical laws, the so-called \emph{mechanistic} models, e.g., models utilizing ordinary and partial differential equations (ODEs and PDEs) or integral equations (IEs).  ML models are usually data-driven models, and the more training data are available, the better are the model-derived outcomes. Scientific computing models typically involve  a small number of parameters to describe the system to predict the system's outcome, whereas ML models may depend on a large number of parameters that have to be tuned by the data available. 

It is difficult to say which approach (mechanistic models or ML models) is better. Both have pros and cons. Mechanistic models do not depend on data availability and are easy to interpret. However, they require that a modeler knows a mechanism underlying the model, which may be elusive. In this case, data-driven non-mechanistic models can give a very accurate prediction directly form data. 

But what if no data are available and the mechanism is known only partially? A good example of such a situation is the COVID-19 pandemic. Very limited data are available at the beginning of the pandemic. On the other hand, we have an ODE-based SIR (Susceptible, Infectious, Recovered) model \cite{kermack1927contribution} that is described using only a few parameters and can be fitted by a small dataset. At the same time, this ODE-based model has components that are subject to high uncertainty and require specific strategies to understand (learn) this uncertainty.

SciML provides a scientifically sound approach to handle uncertainty of a physical or biological model using ML algorithms. Reference \cite{dandekar2020machine} provides an example of a SciML approach to gain insights of the COVID-19 pandemic  using the SIR model,  which in its simplest form  has three ODEs:
\begin{equation*}
    \left\{
                \begin{array}{ll}
                  \displaystyle \frac{dS}{dt} = -\beta\frac{SI}{N} + \gamma I\\ \ \\
                  \displaystyle \frac{dI}{dt} = \beta\frac{SI}{N} - \gamma I\\ \ \\
                  \displaystyle \frac{dR}{dt} = \gamma I.
                \end{array}
              \right.
\end{equation*}
Here, $S(t)$, $I(t)$ and $R(t)$ are the number of susceptible, infected, and recovered subjects at time $t$, respectively. The constants $\beta$ and $\gamma$ are the infection and the recovery rates. The total number of subjects in the population $N = S(t) + I(t) + R(t)$ is regarded as a constant, that is, births and deaths are not taken into account. 
The goal is to study the effect of quarantine. For this purpose, the SIR model is augmented by adding a time-dependent quarantine strength rate term $Q(t)$ and a quarantined population $T(t)$, which is prevented from having any further contact with the susceptible population. Therefore, the system of ODEs takes the form:. 
\begin{equation*}
    \left\{
                \begin{array}{ll}
                  \displaystyle \frac{dS}{dt} = -\beta\frac{SI}{N} + \gamma I\\ \ \\
                  \displaystyle \frac{dI}{dt} = \beta\frac{SI}{N} - (\gamma + Q) I\\ \ \\
                  \displaystyle \frac{dR}{dt} = \gamma I + \delta T \\ \ \\
                  \displaystyle \frac{dT}{dt} = Q I - \delta T.
                \end{array}
              \right.
\end{equation*}
Thus, the term $I(t)$ denotes the infected population still having contact with the susceptibles (as done in the standard SIR model), whereas the term $T(t)$ denotes the infected population of subjects who are effectively quarantined and isolated. The constant $\delta$ is an additional recovery rate that quantifies the rate of recovery of the quarantined subjects. Thus, we can write an expression for the quarantined infected population  $T(t)$ as 
$
    T(t) = Q(t) \times I(t).
$
Due to the universal approximation theorem, the quarantine term is replaced by a neural network (NN) and the deterministic system of ODEs is approximated by neural ODEs. 

Thereafter, a neural network was trained by using a small portion of data available, and the developed model was able to predict the infected and recovered counts for highly affected countries in Europe, North America, Asia and South America with a good accuracy. Also, the proposed approach allowed to extract valuable information regarding the efficiency of different quarantine policies \cite{dandekar2020machine}. 

Another interesting and important example of SciML application in the context of COVID-19 pandemic is given in \cite{dandekar2020safe}. It shows how to utilize SciML approach with the data obtained via the contact-tracing apps.  The method presented in the paper is called Safe Blues (https://safeblues.org), and it uses Bluetooth signals similarly to the existing technologies but the method does not require to record information about individuals and their interactions. Instead, it helps to understand population wide dynamics in a privacy-preserving manner. 

There are software packages for implementing SciML, such as the {\em DiffEqFlux.jl} package \citep{rackauckas2019diffeqflux} implemented in the Julia programming language (https://julialang.org).  It   combines the differential equations-based modeling approach with machine learning and neural networks algorithms, and they collectively  provide ready solutions, such as, neural differential equations \citep{chen2018neural} and universal differential equations \citep{rackauckas2020universal} to support research in SciML.

\section{\bf{High Dimensional Inference (HDI)}}\label{sec3}
 
\textcolor{black}{In the current pandemic, large data sets are increasingly available for data mining, analysis and interpretation.  The same is true in many other health science areas, especially in cancer research, where high throughput genomic measurements are available \citep{nakagawa2018whole,rosenbaum2018genomic}. To properly make inference from  complex data sets, novel statistical models and methods are needed to account for the high dimensionality, including cases when the number of predictors may be much larger than the sample size. This section first briefly reviews modern  methods for analyzing big data before we introduce the SSHDI method in the context of high dimensional inference.  Although, to the best of our knowledge, there are currently no COVID-19 data sets with a large number of predictors, i.e. genetic information, we expect that once an appropriate COVID-19 data becomes available, the proposed methodology will be helpful  for  gaining insights into the many puzzling questions on the pathogenesis of COVID-19.}

For moderate sized data sets, there are well-established  statistical methods when the number of predictors diverges with the  sample size.  Some examples are marginal screening in Genome-Wide Association Studies (GWAS) \cite{wang2005genome,wu2010screen} and penalization/regularization methods for {joint modeling and} variable selection \citep{tibshirani1996regression}.
There are also tree-based methods for feature selection that optimize the information gain or gain ratio when generating the decision trees \cite{kotsiantis2013decision,gainratio}.  To enhance weak learners, such as trees, boosting and re-sampling methods have been proposed, along with XGBoost, random forests, among others \cite{breiman2001random,schapire2003boosting,chen2016xgboost}.  However, efficient methods for drawing  inferences from  large  and complex data sets have somewhat lagged but there is now intense research in high dimensional inference (HDI), where the focus is on assessing the uncertainty measures of model parameters,  finding asymptotic distributions of estimated parameters and deriving significance tests or confidence bands.

In the traditional low-dimensional setting when $n > p$ and $p$ fixed, it is well known that the least squares estimator $\hat{\beta}_{LS} = (\bm{X}^\mathrm{T}\bm{X})^{-1}\bm{X}^\mathrm{T}\bm{Y}$ converges to a normal distribution and  exact  inference through p-values and confidence intervals is possible.  However, when $n < p$, the least squares estimation becomes problematic because the sample covariance matrix $\widehat{\bm{\Sigma}}= \bm{X}^\mathrm{T}\bm{X}/n$ is singular. Such problems have become increasingly relevant in the past two decades when   high-throughput data becomes common. The goal is often to find a parsimonious model to study the response variable when there are many  covariates.

\subsection{Current HDI Methods}

Consider the  homoscedastic linear model:
\begin{equation} \label{m0}
\bm{Y} = \bm{X}\beta^* +\bm{\varepsilon}
\end{equation}
where $\bm{Y}=(y_1,y_2,...,y_n)^\mathrm{T}$ is the $n$-vector of responses; $\bm{X}=(X_1,X_2,...,X_p)$ is the $n \times p$ design matrix where the columns contain $p$ covariate vectors $X_j$'s;
$\beta^* = (\beta^*_1,...,\beta^*_p)^\mathrm{T}$ is the true parameter vector of interest and $\bm{\varepsilon}$ is the random noise vector with $\mathbf{E}(\bm{\varepsilon}) = \mathbf{0}_n$.

The high dimensionality referred to herein  includes, but is not limited to the usual case when ``$p >n$,'' such as when $n=500$ samples with $p=1000$ covariates. Even in a ``$p<n$'' setting with $n=1000$ samples and $p=500$ predictors,  direct applications of classic regression models can lead to ambiguous and meaningless estimations and inferences if the number of predictors $p$ is allowed to increase with the sample size $n$. In other words, the classic inferential results for the \emph{fixed $p$} case would not directly apply to the \emph{diverging $p$} case \citep{portnoy1985asymptotic,niemiro1992asymptotics}.  To solve the joint estimation problem, penalized regressions have been widely used, including LASSO \citep{tibshirani1996regression} and some of its many adaptive variants   \citep{fanli,zou2005regularization,candes2007dantzig,huang2008adaptive,lv2009unified}.
The estimators from penalized regressions are shrunk and thus ``irregular'' as their asymptotics become difficult to track.  There are three directions of current research in high dimensional inference:
\begin{itemize}
    \item {\em De-biased methods} derive the $p$-vector of the coeffcients by correcting the known bias of a sparse estimator, for examples, Low Dimensional Projection Estimator (LDPE) in \cite{zhang2014confidence} and de-sparsified LASSO estimator \citep{javanmard2014confidence,buhlmann2014high,dezeure2015high}. 
    The de-biased estimators of $\beta^*$ are for the joint effects of all $p$ predictors and are shown to be  asymptotically unbiased and normally distributed under some regularity conditions and when $p$ is much larger than $n$. Therefore, the individual $p$-values and confidence intervals of the effects can be derived based on the asymptotics.  However, such approaches have limitations
    \citep{fei2019drawing,fei2021estimation}. As the de-biasing procedure relies on accurate estimation of the $p\times p$ precision matrix of the predictors, which itself is a challenging problem \cite{wang2016precision,loh2018high}, finite sample estimation errors are expected. The optimization procedure also involves excessive number of tuning parameters to achieve the desired theoretical properties.
    
    \item {\em Post-selection inference} focuses on valid inference given a {\em selected} model and can be considered as a twin of HDI. \cite{belloni2014inference} proposed a post double selection procedure for estimation and inference with continued work in \cite{belloni2016post,belloni2018valid}, and \cite{lee2016exact} characterized the distribution of a post-LASSO-selection estimator conditioned on the \emph{selected variables}, but only for linear regressions. The apparent limitation is that the post-selection inference cannot detect or correct any errors already made by the selection. For example, if an important feature/predictor was not selected in the first place, the post-selection inference would not retrive it either.
    
    \item {\em High dimensional testing} solves the HDI problem without estimating the coefficients, but derives test statistics for the hypotheses such as $H_0: \beta^*_j =0$ and $H_0: a^\rT \beta^* = c$, where $a$ can be a $p\times q$ matrix but $\text{rank}(a)$ needs to be fixed and not increasing with $n$ or $p$. \cite{ning2017general} proposed the decorrelated score tests for penalized M-estimators; \cite{fang2017testing} introduced a similar procedure based on proportional hazards model; \cite{zhu2018linear} also proposed a method for testing linear hypothesis in high-dimensional linear models. While simplifying the estimation and inference problem to testing could gain robustness and computational advantages, it also loses important information regarding effect sizes and directions.
\end{itemize}

\subsection{SSHDI Method}

A recent and novel framework solves the HDI problem from a different angle and potentially avoids the related limitations. We have shown such an approach has both theoretical and empirical advantages over existing methods \citep{fei2019drawing,fei2021estimation}.
By using multi-sample splitting and smoothing techniques, the novel method converts the challenging high dimensional estimation problem to a sequence of low dimensional estimations. In each of the lower dimensional estimation, the sample size is sufficiently large for the number of predictors \citep{portnoy1985asymptotic}.  
Algorithm \ref{alg1} describes the base procedure, so-called {one-split} estimator, where we first split the original data into equal halves, then apply a  general variable selection procedure to choose a subset of covariates using one half of the data. Next we use the other half of the data to fit partial regressions iteratively using each covariate and the selected covariates. In other words, each coefficient is estimated jointly with the selected subset, which achieves dimension reduction. We show that when the selected covariates contain the sparse active set, the resulting  coefficient  estimator is unbiased, whether the actual effect $\beta^*_j, \; j=1,2,..,p$ is significant or not.

The estimator based on a single split is highly variable due to the random data split and  the variation in the selection.  To this end, we use the idea of bagging \citep{breiman1996bagging} and multi-sample splitting to reduce the variability and increase the power of detecting signals. As shown in Algorithm \ref{alg2}, by repeating the split and estimation a number of times $B$, and aggregating the one-split estimators, SSHDI, shortened for \emph{Split and Smoothing for High Dimensional Inference}, solves the estimation problem of the whole coefficient vector $\beta^*$ in the assumed model (\ref{m0}) with increased power. We show that each coefficient estimator is asymptotically unbiased and normal.  More importantly, the procedure accounts for the variation in model selection, which is largely neglected in most existing works.
We also highlight that the final estimator $\widehat{\beta}$ is robust to various selection methods, such as, sure independent screening (SIS) \citep{fan2008sure},  or regularized regressions with different penalties, like LASSO and SCAD. We denote such a selection method with tuning parameter $\lambda$ by $\mathcal{S}_{\lambda}$.
We further derive a model-free variance estimator based on non-parametric delta method and sub-sampling properties (Algorithm \ref{alg3}) \citep{efron2014estimation,wager2018estimation}. The variance estimators $\widehat{V}_j^B$'s are asymptotically consistent, and possess satisfying empirical performance when the number of re-samples $B$ is of the same order as the sample size, i.e. $B= O(n)$.

\begin{algorithm}
	\caption{One-split Estimator\label{alg1}}
	\begin{algorithmic}[1]
		\REQUIRE A selection procedure $\mathcal{S}_\lambda$ with tuning parameter $\lambda$
		\INPUT Data $(\bY,\bX)$
		\OUTPUT Coefficient estimator $\widetilde{\beta}$
		\STATE Split the data into equal halves $\bD_1$ and $\bD_2$, with sample sizes $|\bD_1|=\lfloor n/2 \rfloor$, $|\bD_2|=\lceil n/2 \rceil$
		\STATE Apply $\mathcal{S}_\lambda$ on $\bD_2$ to select a subset of important covariates $S\subset [p]$
		\FOR{$j=1,..,p$}
		\STATE \textcolor{black}{On $\bD_1 = (\bY^1,\bX^1)$, let $S_{+j} = \{j\}\cup S$ and let $\widetilde{\beta}^1$  be  the coefficient estimator   obtained from the partial regression of
			$\bY^1$   on $\bX^1_{S_{+j}}$}
		\STATE Define $\widetilde{\beta}_{j} = \left(\widetilde{\beta}^1 \right)_j$, which is the coefficient for covariate $X_j$
		\ENDFOR
		\STATE Define $\widetilde{\beta} = (\widetilde{\beta}_1,\widetilde{\beta}_2,..,\widetilde{\beta}_p)$
	\end{algorithmic}
\end{algorithm}

\begin{algorithm}
	\caption{SSHDI Estimator\label{alg2}}
	\begin{algorithmic}[1]
		\REQUIRE A selection procedure $\mathcal{S}_\lambda$ with tuning parameter $\lambda$
		\INPUT Data $(\bY,\bX)$, number of re-samples $B$
		\OUTPUT Coefficient estimator $\widehat{\beta}$
		\FOR{$b=1,2,..,B$}
		\STATE Run Algorithm \ref{alg1} with random data split \textcolor{black}{$(\bD_1^b, \bD_2^b)$}
		\STATE Denote the output estimator as $\widetilde{\beta}^b= (\widetilde{\beta}_1^b,\widetilde{\beta}_2^b,..,\widetilde{\beta}_p^b)$
		\ENDFOR
		\STATE Define $\widehat{\beta} = (\widehat{\beta}_1,\widehat{\beta}_2,..,\widehat{\beta}_p),\;\mathrm{where}\;
		\widehat{\beta}_j = \frac{1}{B} \sum_{b=1}^{B}  \widetilde{\beta}_j^b$ is the average
	\end{algorithmic}
\end{algorithm}

\begin{algorithm}
	\caption{Model-free Variance Estimator\label{alg3}}
	\begin{algorithmic}[1]
		\INPUT $n, B$, $\widetilde{\beta}^b, b=1,2,..,B$ and $\widehat{\beta}$
		\OUTPUT Variance estimator $\widehat{V}^B_j$ for $\widehat{\beta}_j$, $j=0,1,..,p$
		\STATE \textcolor{black}{For $i=1,2,...,N$ and $b=1,2,..,B$, let $J_{bi}\in\{0,1\}$ be the indicator of the $i^{th}$ observation from the $b^{th}$ sub-sample $\bD_1^b$ in Algorithm 2  and let $J_{\cdot i} = \left(\sum_{b=1}^{B}J_{bi}\right)/B$}.
		\FOR{$j = 0,1,..,p$}
		\STATE Define \begin{equation*}
		\widehat{V}_j= \frac{4(n-1)}{n} \sum_{i=1}^{n}\widehat{\mathrm{cov}}^2_{ij}
		\end{equation*}
		where
		\begin{equation*}
		\widehat{\mathrm{cov}}_{ij}=\frac{1}{B} \sum_{b=1}^{B}\left( J_{bi}- J_{\cdot i}\right) \left( \widetilde{\beta}^b_j-\widehat{\beta}_j \right)   
		\end{equation*}
		\STATE Define \begin{equation*}
		\widehat{V}_j^B = \widehat{V}_j - \frac{n}{B^2}\sum_{b=1}^{B}(\widetilde{\beta}^b_j-\widehat{\beta}_j)^2
		\end{equation*}
		\ENDFOR
		\STATE Set $\widehat{V}^B = \left(\widehat{V}^B_1,\widehat{V}^B_2,\dots, \widehat{V}^B_p \right)$
	\end{algorithmic}
\end{algorithm}

The theoretical properties of the one-split estimator and the SSHDI estimator are available \citep{fei2019drawing,fei2021estimation}. To show the asymptotic consistency and normality of the SSHDI estimators, the selection method $\mathcal{S}_{\lambda}$ has to satisfy the ``sure screening'' property, which requires the selected subsets to pick out the true active set $S^*$ with probability approaching $1$ as the sample size $n$ goes to infinity.  As an example,  LASSO with a proper order of the tuning parameter $\lambda$ \citep{zhao2006model}  and sure independence screening (SIS) with a ``beta-min'' condition \citep{fan2010sure}, among others, satisfy the sure screening property.
Both the one-split estimator and the SSHDI estimator are asymptotically unbiased and normal as the sample size $n$ goes to infinity.  The SSHDI estimator  has a smaller variance because of the bagging effect \cite{buhlmann2002analyzing}.

\subsection{Numerical Studies}
There have been extensive numerical experiments comparing the SSHDI procedure with current methods \cite{fei2019drawing}, and examples with non-linear models in \cite{fei2021estimation}. Here we compare SHDI with two de-biased LASSO estimators, LASSO-Pro \citep{van2014asymptotically} and SSLASSO \citep{javanmard2014confidence}. Under the high dimensional linear model (\ref{m0}), we set $n=200$, $p=500$, the active set $S^*\subset \{1,2,..,p\}$ was a fixed random realization with size $|S^*|=5$, and $\beta^*_{S^*}$ was a fixed realization of $5$ i.i.d. random variables from  $U[0.5,2]$. We consider three correlation structures of the covariate vector $\bx_i$:
\begin{itemize}
	\item Identity: $ \bm{\Sigma}_{p\times p}=\mathbf{I}_{p\times p}$;
	\item Autoregressive (AR(1)): $\bm{\Sigma}_{p\times p}: (\bm{\Sigma})_{jk}=(0.8)^{|j-k|}$;
	\item Compound symmetry (CS) $\bm{\Sigma}_{p\times p}$: \textcolor{black}{$= 0.5\mathbf{I}_{p\times p}+0.5\mathbf{1_p}\mathbf{1_p}^T$ and $\mathbf{1_p}$ is the $p\times 1$ vector of $1'$s}.
\end{itemize}
Table \ref{table1} displays the estimated biases and coverage probabilities, where the coverage probability is defined as the proportion of simulations when the derived $95\%$ confidence intervals cover the true parameter $\beta^*$.
We observe that across the board, SSHDI  gives less biased point estimates for the true signals, and  provides reliable confidence intervals around the nominal level for both  true signals and noise variables. In contrast, both LASSO-Pro and SSLASSO have visible discrepancies in terms of point estimation and inference for the true signals and noise variables.

\subsubsection{Multiple myeloma genomics data}
We analyzed a cancer genomic data with $n=163$ multiple myeloma patients \citep{fei2019drawing}. Our interest  is to detect  association between the $\beta$-2 microglobulin (B2M) and gene expressions. B2M is a continuous variable measuring  a small membrane protein produced by malignant myeloma cells, indicating the severity of disease. 
Identifying genes that are related to B2M is clinically important as it helps construct molecular prognostic tools for early diagnosis of disease.

We used the target gene approach  KEGG \citep{hgu133plus2} to identify gene pathways that were shown to be related to cancer development and progression. There were $p=789$ unique probes from the identified pathways and we took the logarithm of both the B2M test value and the gene expressions, {respectively,} as the response and predictors for model (\ref{m0}). We applied SSHDI with LASSO as the selection method and $B=500$ re-samples were drawn for smoothing.

Table \ref{tab_mm} shows the proposed method offered new biological insights with two significant probes at the 5\% family-wise error rate level after adjusting for the Bonferroni correction: 204171\_at (RPS6KB1) and 202076\_at (BIRC2). The two de-biased LASSO estimators found no significant probes.   Both detected genes are highly associated with malignant tumor cells: RPS6KB1, member of the ribosomal protein S6 kinase (RPS6K) family, altercation/mutation has been related to numerous types of cancer including breast cancer, colon cancer, non-small-cell lung cancer, and prostate cancer \citep{sinclair200317q23,slattery2011genetic,zhang2013prognostic,cai2015mir}; BIRC2, whose encoded protein is a member of inhibitors of apoptotic proteins (IAPs) that inhibits apoptosis by binding to tumor necrosis factor receptor-associated factors TRAF1 and TRAF2 \citep{saleem2013inhibitors}, has been related to lung cancer and lymphoma \citep{wang2010overexpression,rahal2014pharmacological}.

\begin{table*}
	\caption[caption]{Comparisons of SSHDI with LASSO-Pro and SSLASSO.
		Rows consist of	$5$ true signals and the average of noise variables.
		In each cell, top number is for SSHDI; middle number is for LASSO-Pro; lower number	is for SSLASSO. \label{table1}}
	\centering
	\resizebox*{\textwidth}{!}{
		\begin{tabular}{ c ccc cc cc}
			\toprule
			& &\multicolumn{2}{c}{Identity} &\multicolumn{2}{c}{AR(1)}&\multicolumn{2}{c}{CS}\\
			\midrule
			Index& $\beta^*_j$& Bias ($\times 10^{-3}$) & Cov Prob (\%) & Bias ($\times 10^{-3}$)& Cov Prob (\%) & Bias ($\times 10^{-3}$) & Cov Prob (\%)\\
			\midrule
			78 & 1.07 & \shortstack[r]{-1.77\\ -81.78 \\ -79.33} & \shortstack[r]{90.5\\ 70.5\\ 90.5} &  \shortstack[r]{10.43 \\ -44.09 \\ -101.95} & \shortstack[r]{92.5 \\ 86 \\ 84.5}  &  \shortstack[r]{-0.35 \\ -38.43 \\ -113.72} & \shortstack[r]{96.5 \\ 92.5 \\ 92.5} \\
			\midrule
			102 & 1.04 & \shortstack[r]{-1.04 \\ -80.28 \\ -77.72 }& \shortstack[r]{96.5\\ 76 \\ 93.5} &  \shortstack[r]{9.70 \\ -44.54 \\ -99.66} &\shortstack[r]{ 92 \\ 87 \\ 82}&  \shortstack[r]{2.44 \\ -32.42 \\ -105.60} & \shortstack[r]{95 \\ 89 \\ 92} \\
			\midrule
			242 & 1.19 & \shortstack[r]{-1.62 \\ -89.43\\ -88.69} & \shortstack[r]{94 \\ 71.5 \\ 87.5 }&  \shortstack[r]{15.58 \\ -47.57 \\ -104.25} & \shortstack[r]{93.5 \\ 88.5 \\ 84 }  &  \shortstack[r]{-4.67 \\ -40.39 \\ -115.51 } & \shortstack[r]{96.5 \\ 91.5 \\ 92 }\\
			\midrule
			359 & 1.43 & \shortstack[r]{-0.14 \\ -75.87 \\ -80.91} & \shortstack[r]{94 \\ 81 \\ 94} & \shortstack[r]{2.98 \\ -41.40 \\ -98.14} &\shortstack[r]{ 96.5 \\ 88 \\ 85} &  \shortstack[r]{2.01 \\ -30.61 \\ -107.5 } & \shortstack[r]{95 \\ 91 \\ 89} \\
			\midrule
			380 & 0.62 &\shortstack[r]{-3.57 \\-84.86\\ -85.73 }& \shortstack[r]{95.5 \\ 75 \\ 89.5} & \shortstack[r]{ 0.54 \\ -60.80 \\ -111.11} & \shortstack[r]{93 \\ 88 \\ 81.5 } &  \shortstack[r]{5.88 \\ -24.20 \\ -99.26} & \shortstack[r]{91.5 \\ 86.5 \\ 90.5 }  \\
			\midrule
			- & 0 & \shortstack[r]{-0.46 \\-0.40 \\-0.27} &\shortstack[r]{ 95\\ 97 \\ 99.5 }& \shortstack[r]{0.65 \\ 3.16 \\ 4.15}& \shortstack[r]{94.82 \\ 96.46 \\ 99.69} & \shortstack[r]{3.26 \\ 5.24 \\ 26.88 }& \shortstack[r]{95.16 \\ 96.34 \\ 99.94  }  \\
			\bottomrule				
	\end{tabular}  }
\end{table*}

\begin{table}
	\centering
	\caption{Top 6 most significant and bottom 6 least significant genes from SSHDI on Multiple Myeloma genomic data. $p$-values are after Bonferroni adjustment.}\label{tab_mm}
	\begin{tabular}{lrrr}
		\hline
		Gene & $\widehat{\beta}$ & SE & $p$ \\
		\hline
		204171\_at (RPS6KB1) & -0.20 & 0.042 &  0.002 \\
		202076\_at (BIRC2)& -0.17 & 0.041 &  0.037 \\
		220414\_at & -0.20 & 0.05 & 0.14 \\
		220394\_at & -0.18 & 0.05 & 0.59 \\
		206493\_at & -0.19 & 0.06 & 0.63 \\
		209878\_s\_at & -0.17 & 0.05 & 0.69 \\
		...&&&\\
		207924\_x\_at & $5\times 10^{-4}$ & 0.07 & 1 \\
		205289\_at & $-4.4\times 10^{-4}$ & 0.06 & 1 \\
		203591\_s\_at & $4.7\times 10^{-4}$ & 0.07 & 1 \\
		224229\_s\_at & $2.4\times 10^{-4}$ & 0.06 & 1 \\
		217576\_x\_at & $2.5\times 10^{-4}$ & 0.07 & 1 \\
		201656\_at & $2.5\times 10^{-4}$  & 0.08 & 1 \\
		\hline
	\end{tabular}
\end{table}	

\subsection{{Extensions}\label{sec3.4}}
The SSHDI method takes advantage of the multi sample-splitting in \cite{meinshausen2009p} and the bagging idea in \cite{buhlmann2002analyzing}.  It is thus fundamentally different from methods based on penalized regressions for high dimensional predictors.  As the data split separates selection and estimation, the SSHDI estimator and inferences are not sensitive to the tuning parameters used for variable/model selection, which is a major drawback of the current methods \citep{buhlmann2014high,ning2017general}. Furthermore, the variance estimator is free of the parametric model and achieves variance reduction from the effect of bagging. 

 There are clear computational advantages of the proposed procedure. First, the concise  algorithms  are justified by theory and straightforward to implement in real data applications. Second, as SSHDI uses multiple data-splitting, it is naturally suitable for parallel computing to greatly speed up the computing time.  In particular, the procedure can be paralleled for both the number of splits $B$ and the partial regressions iterating among $p$ covariates, thus taking full advantage of the multi-core CPU or GPU computing. On the other hand, SSHDI performance might scale up with the sample size, since the number of re-samples required is $B = O(n)$. The estimation and inference accuracy also depends on the quality of the model selection results, where the sure screening property is crucial. Further technical details are available in \citep{fei2019drawing,fei2021estimation} and in a new application that further demonstrates the flexibility of the procedure to make accurate inference from complex data in a different type of medical trial.

Specifically, \citep{fei2021estimation} applied the SSHDI procedure to a lung cancer study with high dimensional genetics data. The goal was to find significant predictors among $13,663$ SNPs and SNP-smoking interactions that are associated with lung cancer patients in a case-control study with sample size $N=1,459$. SSHDI was extended from linear regression to generalized linear models (GLM), and comparison between SSHDI and LASSO-Pro showed that the former was able to identify more significant predictors than the latter (9 versus 2) and with faster computation time.

Because the core of the SSHDI procedure is an aggregation method that involves re-sampling, averaging of base learners, and model-free inferences, it has many applications and extensions. For example, it is straightforward to extend the framework to generalized linear models \citep{fei2021estimation}, survival models, mixed effects models, among others. Similar ideas for inferences have also been studied in the context of random forests and predictions \citep{wager2018estimation}.  The idea of split and smoothing can also  be applied to problems beyond regression models. It can be used for dimension reduction and tackle estimation problems without resorting to penalization but still have desired properties. In addition, the model-free variance estimation and inference approaches add extra flexibility to the framework.
 
There are recent research studies on understanding the disease and genome sequencing of the SARS-CoV-2 virus \citep{forster2020phylogenetic,zhang2020genomic}, including a comprehensive data collection effort on the population level for COVID-19 cases and deaths \citep{covid19cdc}.  There is now a  data gathering and repository \citep{hamzah2020coronatracker} that helps to model trends of the spread and make predictions of COVID-19 \citep{petropoulos2020forecasting}. Recently, \cite{chang2} proposed a hierarchical agglomerative algorithm for pooled testing with a social graph that could lead to roughly $20\%-35\%$ cost reduction compared to random pooling by using the Dorfman two-stage method when samples within a group are positively correlated. As large scale  phenotype and genotype data at the individual level become available in this  repository and other sources,  we expect HDI analysis to play an important role in understanding the COVID-19 pathology. For example, HDI models will be able to estimate and test the significance of risk factors related to health outcomes in the presence of many confounders. 
We also see potential extensions of the SSHDI method to prediction and learning problems with a large number of features.

\section{\bf \textcolor{black}{Computational Epidemiology Advances}}
\label{sec4}
Recent years have seen a proliferation of healthcare-related data inventory and cloud-driven software that are used to solve computationally challenging data science problems, especially those related to computational epidemiology. One example is infection source detection (i.e., searching for superspreaders during the COVID-19 pandemic) or to enforce quarantine measures efficiently, which finds applications in {\it digital contact tracing} that employs human contact tracers or mobile technologies (e.g., wireless Bluetooth to measure social connectivity) to trace the social contacts of infected person as well as searching for the outbreak origin \cite{contacttracing1,contacttracing2}. These computational epidemiology problems typically involve networks that arise from social relationship and mobility in the network and require algorithmic advances for computational speedup.

Epidemic spreading patterns can be discovered by knowing who infects whom in an outbreak by modeling users and social contacts between users as vertices and edges respectively. However, digital contact tracing in a viral outbreak may lead to huge graphs whose veracity, volume and size may impede a direct use of standard graph algorithms. For example, running standard graph algorithms such as the breadth-first-search algorithm for every single vertex of a massive graph can become computationally impractical. Scalable cloud computing or machine learning can possibly alleviate this data challenge to some extent. 

In addition, computational epidemiology problems often have statistical features that cannot be easily modeled mathematically. For example, the social network topology in digital contact tracing may have missing or noisy information and runs the risk of being out-dated. The data may also have local and global statistical dependencies that affect the problem-solving approach and its solution quality. {\it In general, the inherent statistics of data influences algorithmic tuning and consequentially computational performance.} Designing computational techniques that exploit statistical features is thus key to algorithmic speedup without incurring significant information loss or degraded solution quality. We give an overview of {\it Network Centrality as Statistical Inference} \citep{shah,cheewai1,cheewai2,cheewai3} and Graph Neural Networks as useful frameworks to design scalable algorithms for digital contact tracing and other computational epidemiology optimization problems.

Consider the problem of tracing infection source in \cite{shah,cheewai2,cheewai3}: Given a snapshot observation of the social graph with the infected users, who is the Patient Zero that causes the outbreak? Let us model the cascading over a graph $G$ by the susceptible-infectious (SI) model in the epidemiology literature. The SI model assumes that a user once infected possesses the disease and in turn spreads the disease to one of his or her susceptible neighbors. A snapshot observation of the cascade is  $G_n$, where n is the number of infected users (modelled by the vertices in the graph $G_n$, which is a subgraph of $G$). For a given social graph $G_n$ over the underlying graph $G, v^*$ is a maximum likelihood estimator for the source in $G_n$, i.e., $ P(v^*|G_n) = \max_{v_i \in G_n} P(v_i|G_n)$. By Bayes theorem, $P(G_n|v)$ is the probability that $v$ is the actual Patient Zero whose initial infection leads to the social graph data $G_n$. Now, let $\sigma$ be a possible \textcolor{black}{spreading order (defined as a sequence of distinct vertices sequence starting from $v$ and containing all the infected subgraph vertices)}, and let $M(v,G_n)$ be the collection of all the spreading orders starting from $v$ as the source in $G_n$. The likelihood function is
$$
P (G_n| v) = \Sigma_{\sigma  \in M (v,G_n)}P (\sigma | v).
$$

Given the observation $G_N$, the node that is most likely to be the epidemic source can be obtained by solving the maximum-likelihood estimation problem: \cite{shah,cheewai2,cheewai3}:
\begin{equation}
\hat{v}  \in  \arg \max_{v\in G_N} P (G_N| v).
\label{eq:est}
\end{equation}
 
Both the size and the combinatorial nature of the problem makes solving (\ref{eq:est}) computationally challenging. For instance, when $G_N$ is a general tree graph, (\ref{eq:est}) is still an open problem even when we consider the simplest Susceptible-Infectious (SI) spreading model \cite{shah,cheewai2}. However, if $G$ is an infinite degree-regular tree graph, one can show that $P (\sigma| v)$ for any vertex $v$ is equal, and thus solving (\ref{eq:est}) reduces to simply counting $M(v,G_n)$ (also known as the rumor centrality \cite{shah}). The node with the largest rumour centrality is the rumour center, which is equivalent to the tree centroid in graph theory \cite{cheewai3}. However, when $G$ is finite or $G_n$ is a general graph with cycles, then each $P (\sigma| v)$ in (\ref{eq:est}) is different, and so evaluating the likelihood function requires computing $M(v,G_n)$ and also tracking $P (\sigma| v)$ for all spreading orders, making (\ref{eq:est}) harder to solve. Apart from identifying special cases in which (\ref{eq:est}) can be solved optimally (e.g., degree-regular trees with an underlying infinite graph), it is interesting to solve (\ref{eq:est}) by a network centrality-based approach that allows a graph-theoretic interpretation and retains some of its intuition. 

\subsection{Reverse Engineering Approach}

In  {\it reverse engineering}, we ask: \textcolor{black}{Given a well-known network centrality, what are relevant inference problems  to computational epidemiology that they implicitly solve?}

The appropriate network centrality can succinctly capture the effect of stochastic processes on the graph, and its algorithms can be useful for computing exact or approximate solution to statistical inference optimization problems. For example, the rumour centrality in \cite{shah} is statistically optimal only when the graph is degree-regular, and can serve as a good heuristic to find approximately good solution. Hence, a network centrality perspective can provide guiding principles on algorithm design even when the original problem is hard to solve. The value of reverse engineering thus lies in giving theoretical insights to the solvability  of the problem and whether a solution is near-optimal or not.

\subsection{Forward Engineering Approach}
In  {\it forward engineering}, we ask: Given a stochastic optimization formulation over a network, how to transform it or to decompose it to one whose subproblems are graph-theoretic and can utilize network centrality, then solve or approximate the overall problem? Answering these questions thus entails an algorithmic approach that seeks to simplify the original problem, making the problem-solving methodology scalable to accommodate practical situations and low-complexity data algorithms.

For instance, even though the rumor centrality approach in \cite{shah} is optimal only for graphs that are degree-regular, the fact that the rumor center is equivalent to either the distance center \cite{shah} or the graph centroid in \cite{cheewai3} opens doors to new algorithmic methodology associated with distance centrality or branch weight centrality respectively. This can lead to fast algorithms for processing graphs that are not degree-regular, serving as a good heuristic to solve (\ref{eq:est}) for the general case. In other words, the forward engineering approach enables the reuse of existing algorithms or a performance comparison between different graph algorithms. It also provides a message-passing (i.e., belief propagation) algorithmic perspective to improve existing network centrality-based algorithms \cite{shah,cheewai1,cheewai2,cheewai3}.

Another instance of forward engineering is the problem of minimizing the disease spread, where the vaccine centrality is proposed in \cite{cheewai3} as an approximation algorithm to solve a statistical estimation problem. The approach of using network centrality as a statistical tool for inference can be generalized from a static network to time-dependent networks when real-time data or more accurate spreading models for COVID-19 are available \cite{chang1,poor1,poor2}. Finding the appropriate network centrality to explain flow patterns or temporal scales of changes in the network is of practical importance. There are also connections between network centrality as statistical inference and graph signal processing, which include methods for sampling, filtering or machine learning over graphs. The confluence of these research directions can lead to mathematically rigorous graph analytics for analyzing contact tracing and other computational epidemiology problems in large networks.

\subsection{Graph Neural Network Approach}
In this section, we describe a graph neural network (GNN) learning methodology to solve (\ref{eq:est}). The idea of GNN learning is to encode structured features of the graph data into a neural network by applying recurrent layers to each node of the graph along with some form of approximation as in recent applications of GNN to combinatorial optimization over graphs \citep{khalil2017learning}. As the neural network weights are trained using semi-supervised examples with labels to capture structural properties of all the nodes in the graph input data, GNNs can be leveraged to address problems related to network centrality. An advantage of using GNN lies in complexity reduction. For example, graph algorithms like the Breadth-First Search algorithm have complexity $O(N+|E|)$ where $E$ is the edge set of the graph. To compute a solution to (\ref{eq:est}) with low complexity, one approach is to approximate the spreading order probabilities in the graph instead of keeping track of all possible probabilities. Another approach is dimension reduction by ignoring nodes near to the graph boundary when the graph is sufficiently large so that accurate GNN models can be trained by only nodes of interests in order to reduce computational time.

\begin{figure*}
\centerline{\includegraphics[scale=0.30]{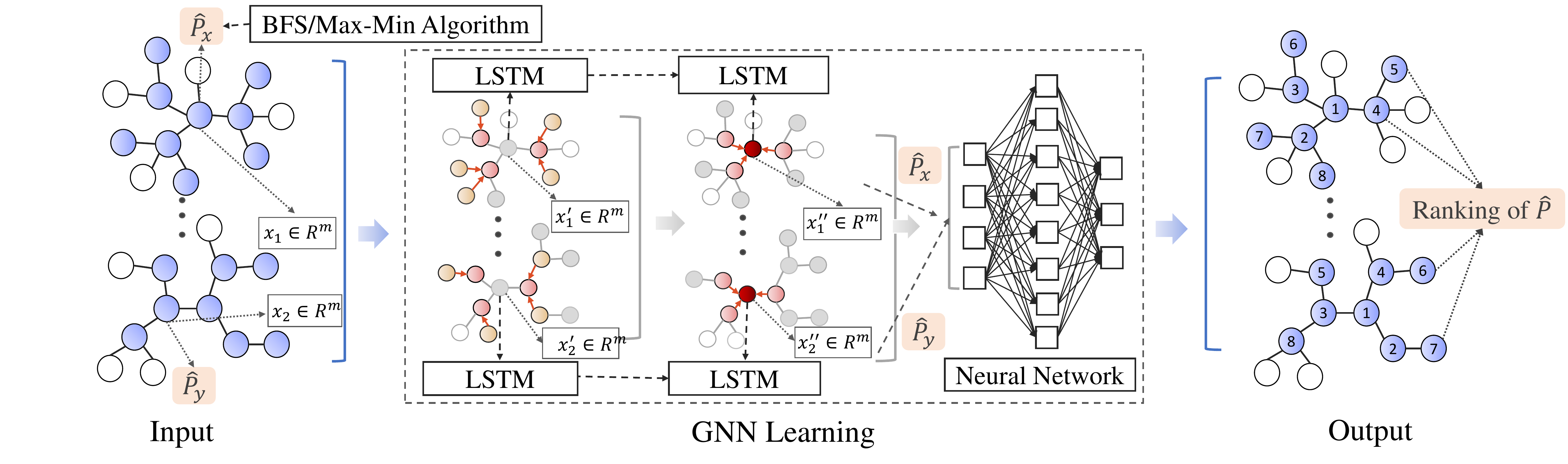}}
\caption{\label{fig:gnn} The overall architecture of an epidemic source inference by node regression using a graph neural network. The input data is a number of smaller contact tracing networks, where each node has a few structural features, labeled with an approximation of the permutation probability. We then use the GraphSage algorithm in \cite{graphsage} with LSTM aggregators as our training model to output a prediction of the spreading order probability for each node as the solution of (\ref{eq:est}) for the input of bigger networks.}
\end{figure*}

The training stage of GNN is important to capture simultaneously the inherent statistical and topological features of a graph data. One possible way is to first generate a training set using a number of graphs that are typically small in size (e.g., hundreds of nodes) and to augment descriptors with the structural features for each node of the graphs as the input data. For these smaller graphs, we solve (\ref{eq:est}) to find approximately the permutation probabilities of these nodes that are then used as the training set labels for the GNN regression as shown by the GNN architecture in Figure \ref{fig:gnn}.

\subsubsection{Node Feature Selection and Labeling}
There are several possibilities to identify useful features and construct labels for each node of the graph input data. Let us consider some basic graph-theoretic features such as the degree and distance. For example, given a snapshot of epidemic network (with the vertex set $\mathcal{V}$) as shown in Figure \ref{fig:example}, some node features can be obtained as follows:

\begin{figure}
\centerline{\includegraphics[scale=0.4]{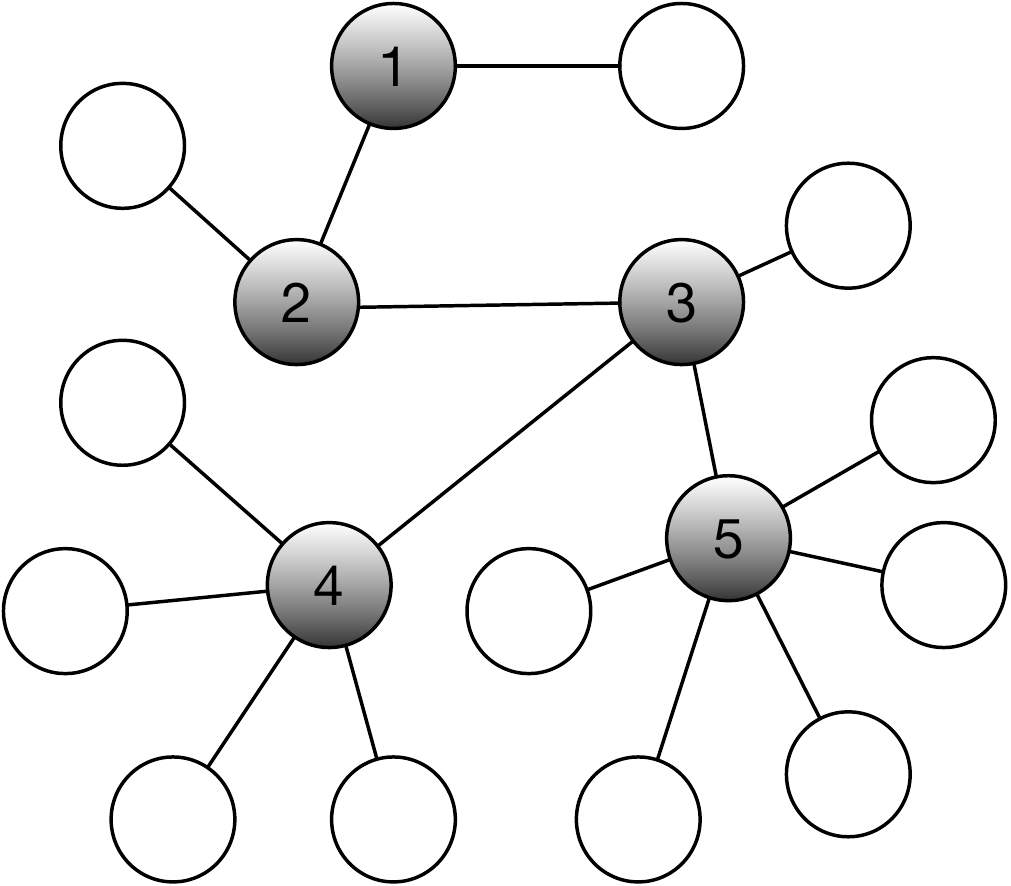}}
\caption{\label{fig:example}An infection network with a degree-irregular tree as the underlying graph, where the degrees of the shaded (infected) nodes $1,2,3,4,5$ are $2,3,4,5,6$, respectively.}
\end{figure}

$\mathbf{Degree \; ratio}$: This is the ratio of the degree of a node $v_i$, say $d(v_i)$, to the sum of the degrees of all the other nodes:
$$r(v_i) = \frac{d(v_i)}{\sum_{v\in \mathcal{V}}d(v)}.$$
For example, the degree rate of nodes $1,2,3,4,$ and $5$ in Fig. \ref{fig:example} are $\frac{1}{16},\frac{3}{32}, \frac{1}{8}, \frac{5}{32} $ and $\frac{3}{16}$, respectively.

 $\mathbf{Infected \; proportion\; ratio}$: This is the ratio of the number of infected neighbors of a node to the sum of the uninfected nodes in the graph:
$$\widehat{r}(v_i) = \frac{\widetilde{d}(v_i)}{\sum_{v\in\mathcal{V}}\widetilde{d}(v_i)}.$$
For example, the infected proportion ratio of nodes $1,2,3,4,$ and $5$ in Fig. \ref{fig:example} is $\frac{1}{12}, \frac{1}{12}, \frac{1}{4}, \frac{1}{12}, $ and $\frac{1}{12}$, respectively.

$\mathbf{Labels}$: The training label of a node is an approximation or averaging of its permutation probabilities obtained by solving (\ref{eq:est}) using any network centrality algorithm or standard graph algorithms (e.g., Breadth-First Search).

\subsubsection{Node regression using Graph Convolutional Network}

Once the feature selection stage is completed, we train the GNN to learn a function by generating the node embedding based on the selected features and the topological structures of each node in the graph data by some form of iterative updates \cite{scarselli2008graph}. At each layer of the neural networks, the vertex $v$ in GNN can be updated as follows:
\begin{equation}
\begin{aligned}
\alpha_v^{(k)} = &\mathrm{Aggregate}^{(k)}(\{\beta_u^{(k-1)}:u\in\mathcal{N}(v)\},\{\gamma_x:x\in\\
&\varepsilon(v)\})\\
\beta_v^{(k)}  = &\mathrm{Combine}^{(k)}(\beta_u^{(k-1)},\alpha_v^{(k)}),
\end{aligned}
\label{eq:gnn1}
\end{equation}
where $\mathcal{N}(v)$ denotes the set of the neighbors of $v$, $\varepsilon(v)$ denotes the set of edges with $v$ as one end node, $\beta_v^{(k)}$ denotes the $k$-th layer's output feature of vertex $v$, and $\alpha_v^{(k)}$ is the aggregate iterate. The learning process can be accomplished by inductive graph neural network training, e.g., GraphSage in \cite{graphsage}, and the following LSTM aggregator:
$$\alpha_v^{(k)} = \mathrm{LSTM}(\{\beta_u^{(k-1)}:u\in\mathcal{N}(v)\}),$$
and the Rectified Linear Unit (ReLU) combination function:
$$\beta_v^{(k)}=\mathrm{max} \left\{ 0, W^{(k)}\cdot \frac{1}{|\mathcal{N}(v)|+1}\sum_{u}\beta_v^{(k-1)}\right\}, 
$$
where $u \in \mathcal{N}(v)\cup\{v\}$ and $\{W^{(k)}\}$ are the weight matrices to be updated. This GNN framework can be extended with more advanced deep learning techniques or integrated with other network centrality-based algorithms. 

\begin{figure*}[htp]
    \centering
    \includegraphics[scale=0.45]{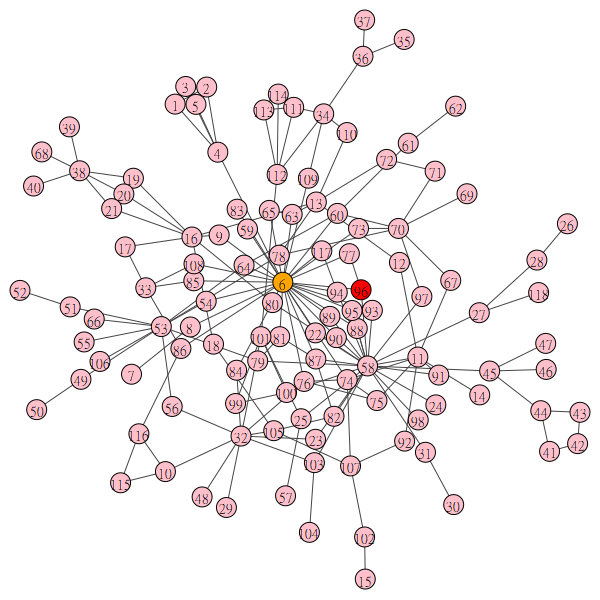}
       \includegraphics[scale=0.32]{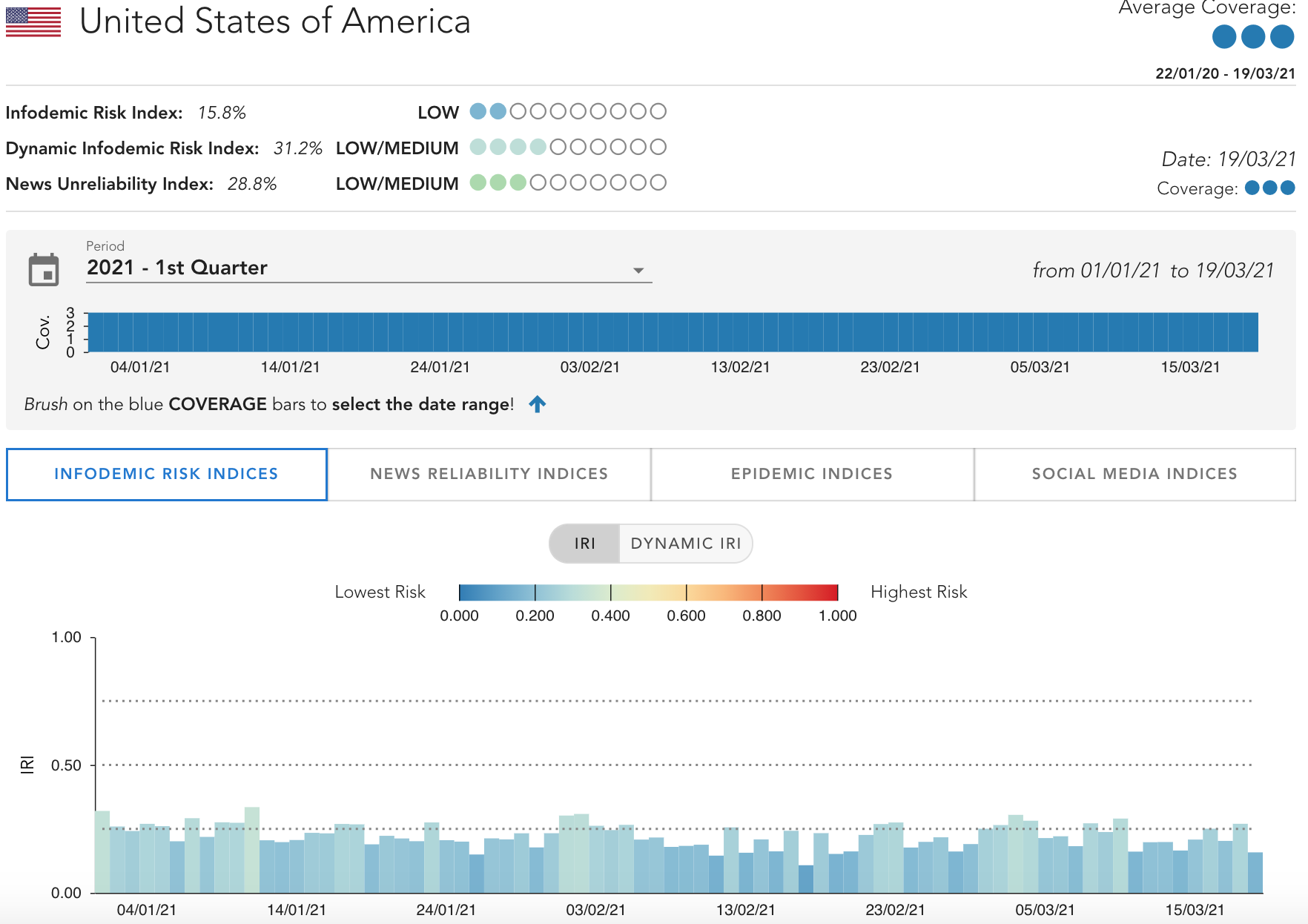}
    \caption{(a) SARS-CoV2003 Contact Tracing Network in Taiwan. Each vertex represents either a confirmed case or a hospital. The orange and red vertices represent the estimated source determined by contact tracing algorithms in \cite{cheewai1} and in \cite{shah} respectively. (b) Photo courtesy of the World Health Organization Infodemic risk management system in \cite{infodemic} to assess the spread of COVID-19 related misinformation.}
    \label{fig:network_prob}
\end{figure*}

We describe briefly a Contact Tracing Algorithm \ref{alggnn} in \cite{cheewai1} that uses a weighted distance centrality measures where the weights are computed for node regression using the aforementioned GNN approach to solve (\ref{eq:est}). As an illustration, by reconstructing the contact tracing network data of SARS-CoV2003 (a virus very similar to the COVID-19 coronavirus) in Taiwan \cite{cheewai1}, this algorithm can correctly identify the first place, Taipei Municipal Heping Hospital (now Taipei City Hospital Heping Branch), of an infection cluster in April 2003 in Taiwan as compared to a breadth-first search heuristic in \cite{shah}, which chooses the red vertex modeling a confirmed index case (not the first case) who had been to the Taipei Municipal Heping Hospital as shown in Figure \ref{fig:network_prob}(a). In addition, the network centrality approach can enable visualization tools (e.g., dot distribution map) to visualize the likelihood of infection source that may be of value to public healthcare policymakers. 

\begin{algorithm}
	\caption{Contact Tracing Algorithm by Ranking Infection Source by Network Centrality and GNN} \label{alggnn}
	\begin{algorithmic}[1]
		\INPUT Infection networks $\{G_1, G_2, ..., G_N\}$ harvested by forward contact tracing 
		\OUTPUT A ranking of the outbreak source probability of each node for backward contact tracing
		\STATE Calculate the structure features (e.g., degree ratio and infected proportion ratio) for each node in the network.
		\STATE Calculate the labels (i.e., approximate solution to (\ref{eq:est})) for each node  according to a network centrality algorithm (e.g., rumor centrality in \cite{shah,cheewai2} or statistical distance centrality algorithm in \cite{cheewai1}).
		\STATE Train the regression model by GraphSage \cite{graphsage} to output a ranking of the probability for each node to be the outbreak source in each network $\{G_1, G_2, ..., G_N\}$.
	\end{algorithmic}
\end{algorithm}

The problem in (\ref{eq:est}) can potentially be useful to other kinds of computational epidemiology problems, such as COVID-19 Infodemic management as introduced by the World Health Organization. In \cite{infodemic}, an Infodemic risk management system, as shown in Figure \ref{fig:network_prob}(b), is developed to visualize and assess the spread of misinformation concerning vaccine and erroneous treatment. Given the volume of COVID-19 related misinformation, finding a rumor source can help limit the damage and spread of false information \cite{infodemic}. In summary, the network centrality as statistical inference and GNN machine learning frameworks are examples of first step towards a theoretically sound and computationally efficient approach to digital contact tracing and computational epidemiology.

\section{\bf{Metaheuristics}}\label{sec5}
This section briefly reviews the increasing role of metaheuristics in big data research.  There are many metaheuristic algorithms and for space consideration, we consider a popular subclass of them called nature-inspired metaheuristic algorithms.  These algorithms are  widely used to tackle high-dimensional and complex optimization  problems in engineering and computer science, and are increasingly used in other disciplines.   They are appealing for several reasons.  They   are general-purpose optimization algorithms, assumptions-free, fast, powerful and easy to implement for solving all kinds of complex optimization problems with hundreds or thousands of variables. Codes for many of them are available in MatLab, and freely on many websites.  Their recent meteoric rise in popularity  in industry and even in academia is nicely documented in \cite{whit1,whit2}.  The promise and excitement of investigating how well metaheuristics performs for tackling problems with millions of variables was the focus of a special issue in Information Sciences \citep{informations}.

Some examples of nature-inspired metaheuristic algorithms that seem more popular  are Particle Swarm optimization (PSO), Differential Evolution (DE), Imperialist Competitive Algorithm (ICA), Competitive Swarm Optimizer (CSO) and Cuckoo search, just to name a few.  Each algorithm has a different motivation from nature and works differently.  A commonality is that each algorithm has a few tuning parameters and a few stochastic elements and each has its own  way of updating its trajectory via a couple of equations    that model a natural phenomenon or an animal's behavior.   For example, PSO mimics a flock of birds fling in the sky and looking for food on the ground.  Each bird has its own idea where the food is (local optimum) but they communicate with one another and collectively  make a decision where the food is on the ground (global optimum) and each bird flies toward it without completely relinquishing  its take where the food is (local optimum).    \cite{Yang} gave a concise description of many such algorithms with illustrative applications and sample codes. However, theoretical properties and  rigorous proofs of convergence for metaheuristic algorithms are generally elusive  but they remain popular because of their widely reported ability to find an optimum or a nearly optimum for all kinds of  optimization problems \citep{whit1,whit2}.  

 In the last decade, medical researchers have increasingly resorted to and continue to use metaheuristic for tackling all kinds of medical problems that involve optimization.  A very common problem  is to use clinical data and cluster patients into various categories of   disease progression  given baseline data.  For example, we want to know in six months, whether a patient will likely have a stable disease, or whether the patient will improve or deteriorate.  Such problems are challenging  because the data set is large, and there is a large number of different types of explanatory variables potentially useful for predicting outcomes accurately.  The problem then translates to selecting a small number of features in the whole data set that best predict the outcome of interest.  For instance, \cite{yushi} predicted disease progression in Idiopathic Pulmonary Fibrosis patients by combining random forest and Quantum Particle Swarm Optimization.  This is an increasingly common optimizing technique where one algorithm is hybridized with another to enhance the search capability by exploiting the particular strengths of the two algorithms so that the hybridized version performs better than each of the individual algorithms. The algorithms involved can be two or more and they can be metaheuristic or deterministic.  A recent application is \citep{hinda}, who showed the Grey Wolf algorithm can be hybridized with PSO for accelerated convergence. A monograph on  hybridized algorithms for enhanced performance with applications is \cite{blum}. 

\subsection{Metaheuristics for the COVID-19 pandemic}\label{sec5.1}
The impact of metaheuristic algorithms, hybridized or not,   can be seen in its increasing use in many sub-specialties in medicine and beyond.  For instance,  in  cardiology research, \citep{cardioica} applied ICA to optimally select a minimum number of features best for diagnosing heart problems.  Similarly,     \cite{cardio1}  used  a modified DE algorithm  and \citep{cardio2} used PSO  and a  Bayesian paradigm to predict heart diseases.  In systems biology, \cite{abdullah} applied PSO to select biological model and estimate parameters in the model, and \cite{parameterest} gave a review of metaheuristics for  estimating parameters.   Likewise, they are also increasingly used in  disciplines that traditionally rely on analytical approaches.  For example, there is notable and recent use of metaheuristics to find  optimal experimental designs for  nonlinear regression models in the biostatistical literature.  Some examples are \cite{lukemire1,xu,zhang,lukemire2}, who respectively applied quantum PSO, DE and a modified CSO algorithm, to search for various types of optimal designs for generalized linear models with several interacting factors and some have random effects. 

Not surprisingly, nature-inspired algorithms have been promptly applied to better understand the various aspects of COVID-19.   Various such algorithms were used and they include PSO, DE and ICA to tackle different aspects of the pandemic.  For example, \cite{hungary} implemented ICA to predict trends in the COVID-19 pandemic in Hungary, \cite{italy} used DE to monitor spread of the COVID-19 virus in Italy, \cite{infect, seir} applied PSO to estimate model parameters  in SEIR models or used PSO to use real time data to estimate and predict death rates caused by COVID-19, and \cite{ctscan} used DE to classify COVID-19 patients from chest CT images.  Most recently, \cite{cva} proposed a COVID-19 optimizer algorithm specifically for modeling and controlling the coronavirus distribution process and one its objectives is to minimize the  number of infected countries to slow down the spread.  The authors also showed their algorithm outperformed PSO and GA by $53\%$ and $59\%$, respectively, and newer created metaheuristic algorithms, like the Volcano Eruption Algorithm and the Grey Wolf optimizer, by $15\%$ and $37\%$, respectively.      

 Pareto Optimization (PO) is a common approach to solve optimization problems with multiple objectives.  \cite{hospital} applied PO to tackle problems posed by  COVID-19, which can infect many people quickly resulting in  huge and sudden requests of medical care at various levels. Coping with how, when and where to admit COVID-19 patients efficiently is a complex multiobjective optimization problem.  For instance,  to decrease the in-bed time, save lives and resources, the choice of the most suitable hospital for the patient  has to be balanced by expected admission time, hospital readiness and severity of the COVID-19 patient.  These are multiobjective optimization problems and the author showed their strategy using data from 254 patients in Saudi Arabia outperformed the lexicographic multiobjective optimization method.   Recently, evolutionary algorithms have made remarkable progress in tackling many types of multiobjective optimization problems \citep{tian4,tian1,tian2} and  we expect metaheuristic algorithms  will  make important contributions to solve COVID-19 multiobjective optimization problems, especially when combined with the latest machine learning advances for tackling COVID-19 problems  \cite{mlcovid,covidigital}.
 
Metaheuristics is not a panacea for optimization problems.  A perennial problem is how to tune the parameters for  accelerated convergence  and ensure that the algorithm converges to the theoretical global optimum.  Both  issues have been active research areas  for a long time; recent advances include \cite{rms,ss}.  Other open healthcare problems in metaheuristics are described in \cite{healthcare}.   
 
\section{\bf{Conclusions}}\label{sec6}
We provide an overview of innovative analytic approaches for gaining insights into big data problems.  We focus on handling healthcare issues relevant to the current pandemic indicated in  Table \ref{tab1} and reiterate that the methodologies are  applicable to other types of big data problems.
 
 There are open challenges in data science for healthcare diagnosis, inference, and pandemic response.  For example, consider the statistical and computational issues for  digital contact tracing and its applications in epidemiology. It is important to have accurate infection spreading models and parameters before robust predictive analytics can be developed to solve the large-scale problems. This means that we have to address challenges for high-fidelity computational algorithms and statistical exploration of the data, where new principles are needed to combine these two aspects. Examples of open issues are: can existing network centrality be reverse-engineered to find optimal estimates for the parameters of most interests in large-scale infection spreading like COVID-19 pandemic? A forward-engineering approach may create new forms of network centrality that possess desirable statistical or computational traits for solving the problem. Can machine learning techniques and massive graph neural networks provide an impetus for breakthrough technologies in analyzing past pandemic behaviors to fight against newly-emerging pandemics? Future research in big data for the health sciences concern three important areas of applications of machine- and deep-learning approaches in modern drug development, namely, adverse event detection, trial recruitment optimization, and clinical drug repurposing, including  big data analytics for various stages in drug discovery and development  \citep{ai-in-dd2020}.

 We close with two remarks.  First,  page limits and the breadth of the field did not allow us to discuss all relevant topics. Some omitted topics include social media text data \citep{chambers2018detecting}, virus lineage \citep{virus}, public health data monitoring \citep{soucie2012public} and analysis of administrative and translation data \citep{jrssa2}. Second,  analysis for big data may alternatively begin with a  selected  subset    of the big data with some optimality properties; see \cite{wang}.  Modern statistical methods are then applied to  infer   the key messages from the subset data to the massive data.  Invariably, the task  to properly analyze big data  is challenging and requires collaboration among statisticians, engineers and computer scientists to jointly  create powerful computing environments, new software, data platforms, data integration systems and state-of-the-art statistical and machine learning techniques. 

\section{\bf{Acknowledgements}}
 The authors are grateful to the three anonymous referees and the associate editor for their thorough review of our manuscript and constructive feedback.

\bibliography{mybib}
\bibliographystyle{IEEEtran}
\begin{IEEEbiography}[{\includegraphics[width=1in,height=1.25in,clip,keepaspectratio]{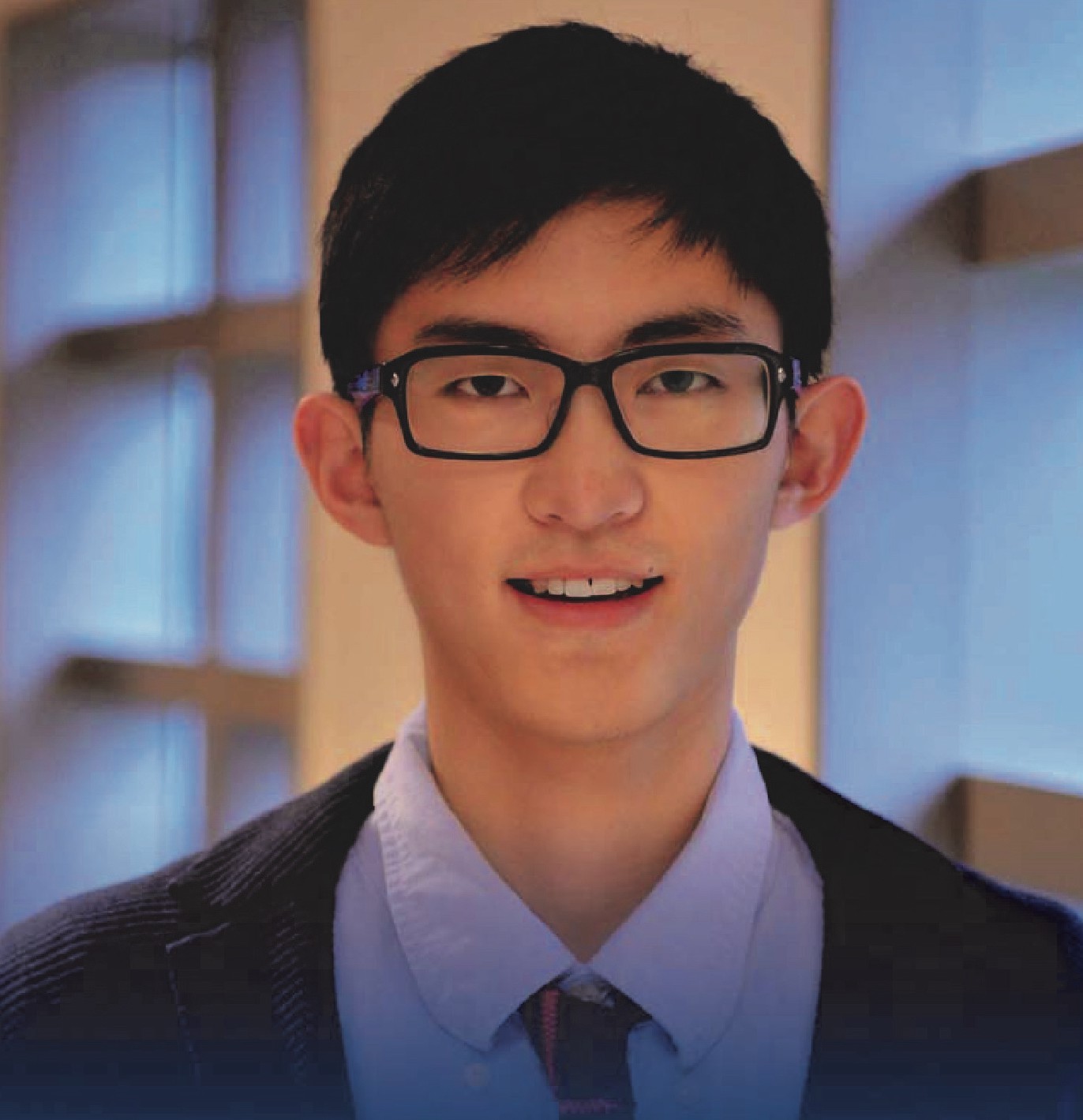}}]{Zhe Fei}
received his PhD in Biostatistics in 2019. He is now an Assistant Professor In-Residence in the Department of Biostatistics at UCLA. His research interests include statistical methods and theories for big data, machine learning and statistical computing, survival analysis, genetics, epigenetics, among others. 
\end{IEEEbiography}

\begin{IEEEbiography}[{\includegraphics[width=1in,height=1.25in,clip,keepaspectratio]{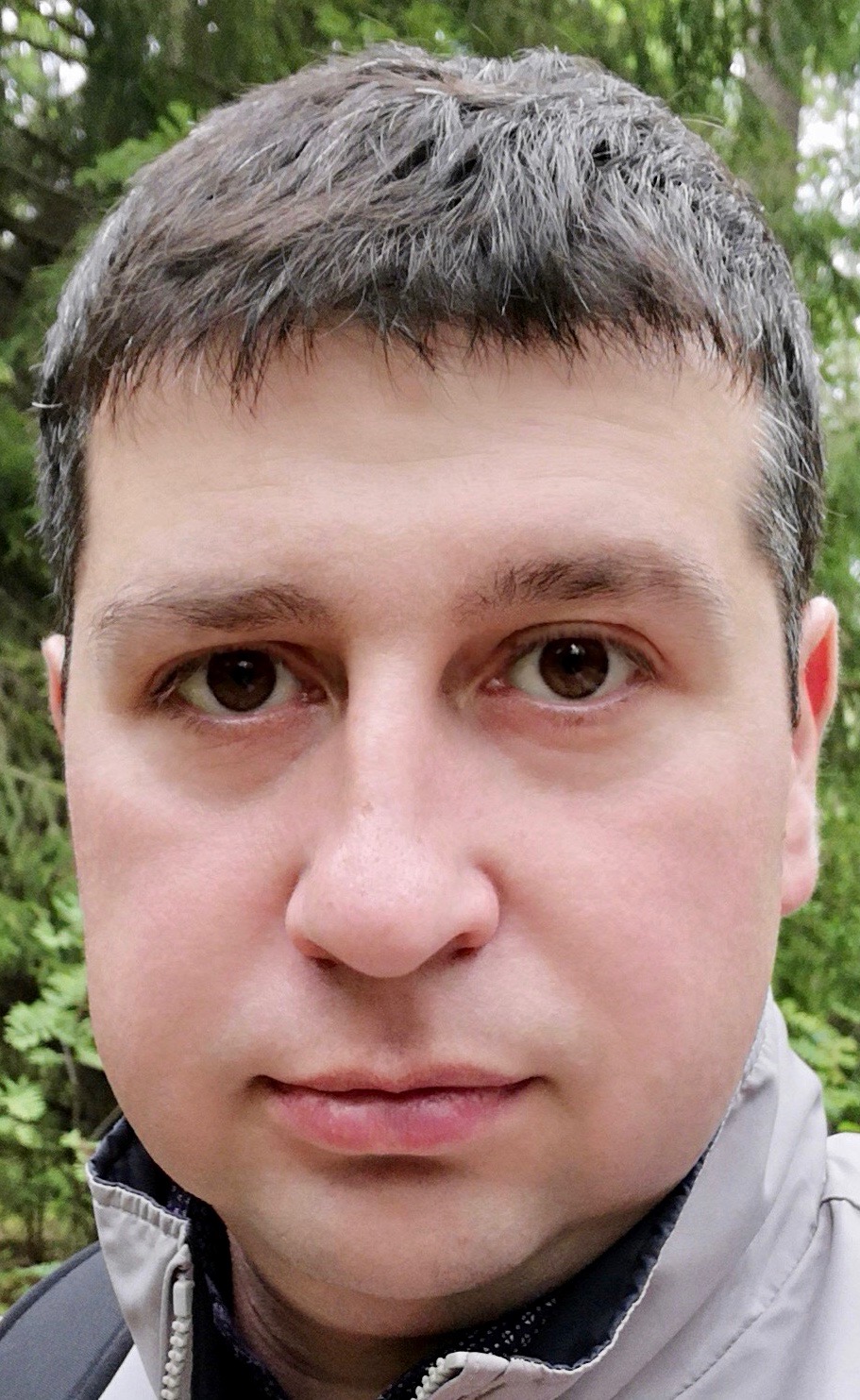}}]{Yevgen Ryeznik}
Dr. Yevgen Ryeznik is senior statistician working in early biometrics and statistical innovation group at AstraZeneca. His research interests include biostatistics, pharmacometrics, machine learning, integral and differential equations and their applications. Recently, he designed and taught two Master/PhD level courses at Uppsala University – on optimal designs and on innovative clinical trials (https://yevgenryeznik.github.io/personal-page/teaching.html)
\end{IEEEbiography}

\begin{IEEEbiography}[{\includegraphics[width=1in,height=1.25in,clip,keepaspectratio]{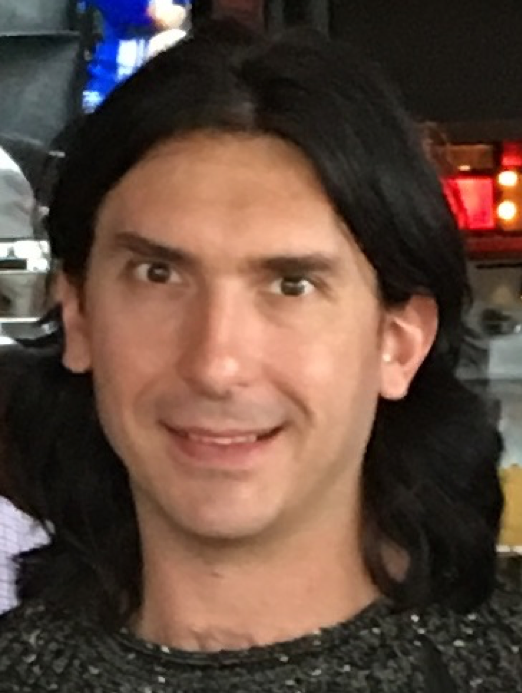}}]{Oleksandr Sverdlov}
Dr. Oleksandr Sverdlov is neuroscience disease area lead statistician in early clinical development at Novartis. He received his PhD in information technology with concentration in statistical science from George Mason University in 2007. He has been actively involved in methodological research and applications on clinical trials in drug development. His most recent work involves design and analysis of proof-of-endpoint clinical studies evaluating novel digital technologies.
\end{IEEEbiography}

\begin{IEEEbiography}[{\includegraphics[width=1in,height=1.25in,clip,keepaspectratio]{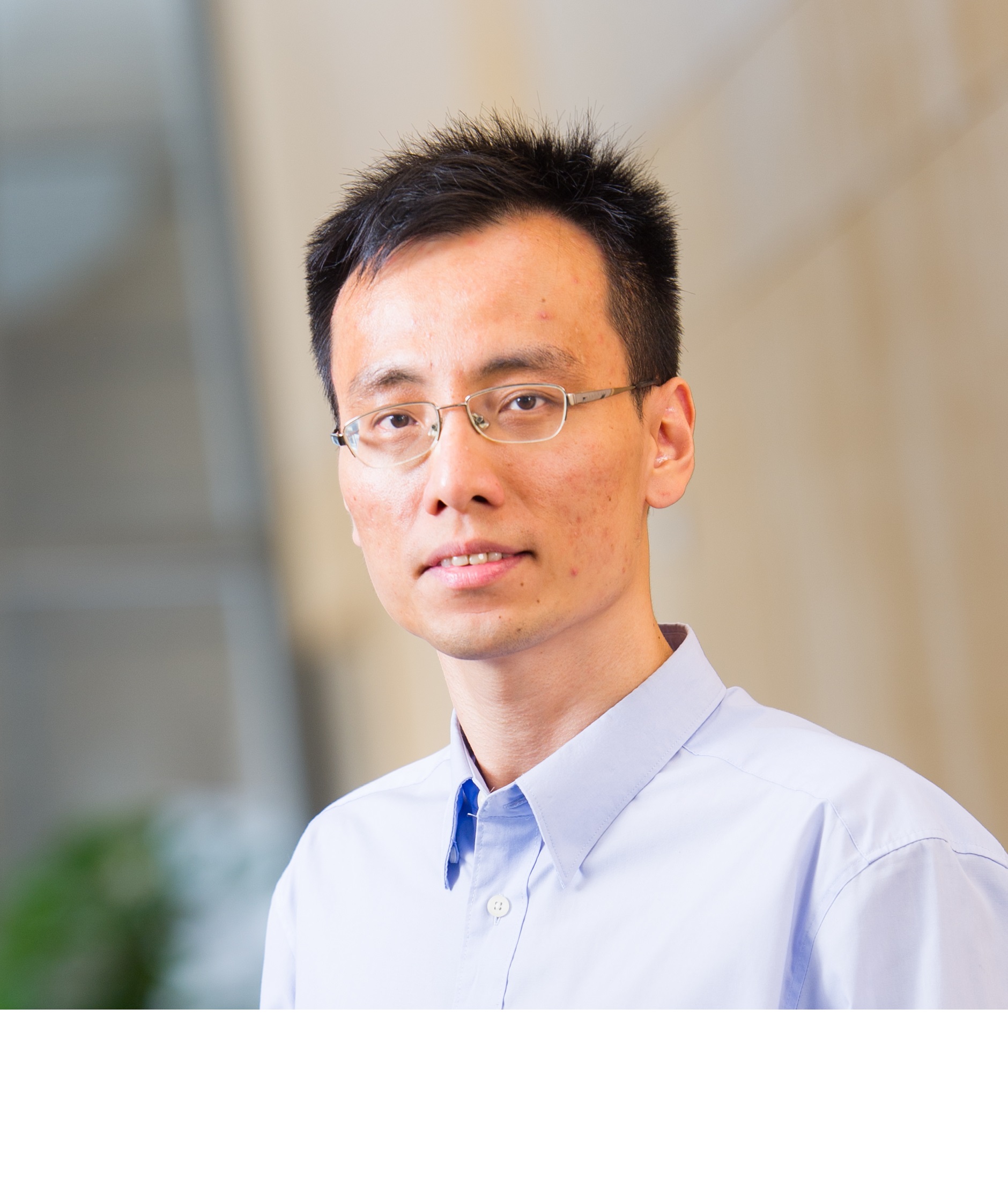}}]{Chee Wei Tan} received his Ph.D. degree from Princeton University, and was a Senior Fellow with the Institute for Pure and Applied Mathematics for the program on Science at Extreme Scales: Where Big Data Meets Large-Scale Computing. His research interests include artificial intelligence, networks, data science and convex optimization theory. He has served as an Editor for the IEEE/ACM Transactions on Networking.
\end{IEEEbiography}

\begin{IEEEbiography}[{\includegraphics[width=1in,height=1.25in,clip,keepaspectratio]{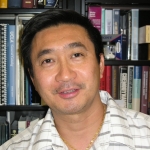}}]{Weng Kee Wong}
Dr. Wong is a Professor of Biostatistics at UCLA, a Fellow of the Institute of Mathematical Statistics, a Fellow of the American Statistical Association and a Fellow of the American Association for the Advancement of Science.  He received his PhD in statistics  from the University of Minnesota and his current research area is in the applications of nature-inspired  metaheuristic algorithms to solve complex  design problems in the biomedical arena.
\end{IEEEbiography}




\end{document}